\documentclass[11pt,a4paper]{article}
\usepackage[hyperref]{emnlp2020}
\usepackage{tikz}
\usepackage{times}
\usepackage{amsmath,graphicx}
\usepackage{latexsym}
\usepackage{CJKutf8}
\usepackage[T1]{fontenc}
\usepackage{multirow}
\usepackage{amssymb}
\usepackage{subfig, graphicx}
\usepackage{url}
\usepackage{dialogue}
\aclfinalcopy % Uncomment this line for the final submission

\definecolor{Seeblue}{rgb}{0.18, 0.51, 0.85}
\definecolor{Gray}{rgb}{0.2, 0.2, 0.2}

\usepackage{microtype}

\title{XPersona: Evaluating Multilingual Personalized Chatbot}
\author{Zhaojiang Lin\thanks{$^*$ Equal contributions. Listing order is random} , Zihan Liu$^*$, Genta Indra Winata$^*$, Samuel Cahyawijaya$^*$,  Andrea Madotto$^*$, \\ \textbf{Yejin Bang, Etsuko Ishii, Pascale Fung} \\
Center for Artificial Intelligence Research (CAiRE)\\
  Department of Electronic and Computer Engineering\\
  The Hong Kong University of Science and Technology, Clear Water Bay, Hong Kong\\
  \texttt{\{zlinao,zliucr,giwinata,scahyawijaya,amadotto\}@connect.ust.hk},\\ \texttt{ pascale@ece.ust.hk}}
  
\date{}

\begin{document}
\maketitle
\begin{abstract}
Personalized dialogue systems are an essential step toward better human-machine interaction. 
Existing personalized dialogue agents rely on properly designed conversational datasets, which are mostly monolingual (e.g., English), which greatly limits the usage of conversational agents in other languages. 
In this paper, we propose a multi-lingual extension of Persona-Chat~\cite{personachat}, namely XPersona.
Our dataset includes persona conversations in six different languages other than English for building and evaluating multilingual personalized agents. 
We experiment with both multilingual and cross-lingual trained baselines, and evaluate them against monolingual and translation-pipeline models using both automatic and human evaluation.
Experimental results show that the multilingual trained models outperform the translation-pipeline and that they are on par with the monolingual models, with the advantage of having a single model across multiple languages. 
On the other hand, the state-of-the-art cross-lingual trained models achieve inferior performance to the other models, showing that cross-lingual conversation modeling is a challenging task. We hope that our dataset and baselines~\footnote{Datasets and all the baselines are available in \url{https://github.com/HLTCHKUST/Xpersona}} will accelerate research in multilingual dialogue systems.

% In this paper, we present XPersona, an extension of Persona-Chat~\cite{personachat}.  than multilingual trained models which indicates 
\end{abstract}

\begin{table*}[t]
\resizebox{0.99\textwidth}{!}{
\begin{tabular}{c|l}
\hline
\multicolumn{1}{l|}{\textbf{Persona}} & I helped design the game starcraft.  I am a famous twitch streamer.. \\ \hline
\textbf{En} & \begin{tabular}[c]{@{}l@{}}\textbf{\textcolor{Gray}{Sys: }}do you know the game starcraft ? i helped designing it !\\ \textbf{\textcolor{Seeblue}{Usr: }}that s awesome , i m sure my ex boyfriend would like that game\\ \textbf{\textcolor{Gray}{Sys: }}i usually stream the game on my twitch channel
that s great , i also love twitch as well\end{tabular} \\ \hline
\textbf{It} & \begin{tabular}[c]{@{}l@{}}\textbf{\textcolor{Gray}{Sys: }}conosci il gioco starcraft? ho aiutato a progettarlo!\\ \textbf{\textcolor{Seeblue}{Usr: }}è fantastico, sono sicuro che al mio ex ragazzo gli piacerebbe\\ \textbf{\textcolor{Gray}{Sys: }}di solito faccio streaming mentre gioco sul mio canale Twitch\end{tabular} \\ \hline
\textbf{Fr} & \begin{tabular}[c]{@{}l@{}}\textbf{\textcolor{Gray}{Sys: }}connaissez-vous le jeu starcraft? j'ai aidé à le concevoir!\\ \textbf{\textcolor{Seeblue}{Usr: }}c'est génial, je suis sûr que mon ex petit ami aimerait ce jeu\\ \textbf{\textcolor{Gray}{Sys: }}Je diffuse généralement le jeu sur ma chaîne Twitch\end{tabular} \\ \hline
\textbf{Id} & \begin{tabular}[c]{@{}l@{}}\textbf{\textcolor{Gray}{Sys: }}apakah anda tahu game starcraft? saya yang mendesainnya!\\ \textbf{\textcolor{Seeblue}{Usr: }}itu luar biasa, saya yakin mantan pacar saya suka game itu.\\ \textbf{\textcolor{Gray}{Sys: }}saya biasanya memainkan game itu di channel twitch saya.\end{tabular} \\ \hline
\textbf{Zh} & \begin{tabular}[c]{@{}l@{}}\textbf{\textcolor{Gray}{Sys: }}\begin{CJK*}{UTF8}{gbsn}你知道游戏《星际争霸》吗？我帮忙设计了它！\end{CJK*}\\ \textbf{\textcolor{Seeblue}{Usr: }}\begin{CJK*}{UTF8}{gbsn}好厉害，我觉得我的前男友会喜欢那个游戏\end{CJK*}\\ \textbf{\textcolor{Gray}{Sys: }}\begin{CJK*}{UTF8}{gbsn} 我经常在我的直播频道上直播游戏\end{CJK*}\end{tabular} \\ \hline
\textbf{Ko} & \begin{tabular}[c]{@{}l@{}}\textbf{\textcolor{Gray}{Sys: }}\begin{CJK*}{UTF8}{}\CJKfamily{mj}\text{너 } \text{게임 } \text{스타크래프트를  아니? } \text{나는 } \text{그것을 } \text{디자인하는 } \text{것을 } \text{도왔어!}\end{CJK*}\\ \textbf{\textcolor{Seeblue}{Usr: }}\begin{CJK*}{UTF8}{}\CJKfamily{mj}\text{멋진데, } \text{내 } \text{전 } \text{남자친구가 } \text{그 } \text{게임을 } \text{좋아할 } \text{거라고 } \text{확신해.} \end{CJK*}\\ \textbf{\textcolor{Gray}{Sys: }}\begin{CJK*}{UTF8}{}\CJKfamily{mj}\text{나는 } \text{보통 } \text{내 } \text{트위치 } \text{채널로 } \text{그 } \text{게임을 } \text{스트리밍해.} \end{CJK*}\end{tabular} \\ \hline
\textbf{Jp} & \begin{tabular}[c]{@{}l@{}}\textbf{\textcolor{Gray}{Sys: }}\begin{CJK*}{UTF8}{min}ゲームのスタークラフトを知っていますか？私はそれを設計するのを助けました！\end{CJK*}\\ \textbf{\textcolor{Seeblue}{Usr: }}\begin{CJK*}{UTF8}{min}それはすごいです、私は私の元彼がそのゲームを好きになると確信しています\end{CJK*}\\ \textbf{\textcolor{Gray}{Sys: }}\begin{CJK*}{UTF8}{min}私は通常、twitchチャンネルでゲームをストリーミングします\end{CJK*}\end{tabular} \\ \hline
\end{tabular}
}
\caption{Multi-turn annotated dialogue samples from test set in seven languages. For simplicity, we only show three turns for each dialogue and the persona in English.}
\end{table*}
\section{Introduction}
Personalized dialogue agents have been shown efficient in conducting human-like conversation. This progress has been catalyzed thanks to existing conversational dataset such as Persona-chat~\cite{personachat,dinan2019second}.
However, the training data are provided in a single language (e.g., English), and thus the resulting systems can perform conversations only in the training language. For wide, commercial dialogue systems are required to handle a large number of languages since the smart home devices market is increasingly international~\cite{etherington_2019}. Therefore, creating multilingual conversational benchmarks is essential, yet challenging since it is costly to perform human annotation of data in all languages. 

A possible solution is to use translation systems before and after the model inference, a two-step translation from any language to English and from English to any language. This comes with three major problems: 1) amplification of translation errors since the current dialogue systems are far from perfect, especially with noisy input; 2) the three-stage pipeline system is significantly slower in terms of inference speed; and 3) high translation costs since the current state-of-the-art models, especially in low resources languages, are only available using costly APIs. 

In this paper, we analyze two possible workarounds to alleviate the aforementioned challenges. The first is to build a cross-lingual transferable system by aligning cross-lingual representations, as in ~\citet{conneau2018xnli}, in which the system is trained on one language and zero-shot to another language. The second is to learn a multilingual system directly from noisy multilingual data (e.g., translated data), thus getting rid of the translation system dependence at inference time. 

To evaluate the aforementioned systems, we propose a dataset called Multilingual Persona-Chat, or XPersona, by extending the Persona-Chat corpora~\cite{dinan2019second} to six languages: Chinese, French, Indonesian, Italian, Korean, and Japanese. In XPersona, the training sets are automatically translated using translation APIs with several human-in-the-loop passes of mistake correction. In contrast, the validation and test sets are annotated by human experts to facilitate both automatic and human evaluations in multiple languages. 

Furthermore, we propose competitive baselines in two training settings, namely, cross-lingual and multilingual, and compare them with translation pipeline models. Our baselines leverage pre-trained cross-lingual~\cite{chi2019cross} and multilingual~\cite{devlin2018bert} models.

An extensive automatic and human evaluation~\cite{li2019acute} of our models shows that a multilingual system is able to outperform strong translation-based models and on par with or even improve the monolingual model. The cross-lingual performance is still lower than other models, which indicates that cross-lingual conversation modeling is very challenging. The main contribution of this paper are summarized as follows:

\begin{itemize}
    \item We present the first multilingual non-goal-oriented dialogue benchmark for evaluating multilingual generative chatbots.
    \item We provide both cross-lingual and multilingual baselines and discuss their limitations to inspire future research.
    \item We show the potential of multilingual systems to understand the mixed language dialogue context and generate coherent responses.
\end{itemize}

\begin{table*}[t]
\centering
\resizebox{0.76\textwidth}{!}{
\begin{tabular}{r|c|c|c|c|c|c|c|c}
\hline
\multicolumn{1}{l|}{} & \multicolumn{4}{c|}{\textbf{Valid.}} & \multicolumn{4}{c}{\textbf{Test}} \\ \hline
\multicolumn{1}{c|}{\textit{\textbf{Lang}}} & \textit{\textbf{\#Dial.}} & \textit{\textbf{\#Utt.}} & \textit{\textbf{Edit}} & \textit{\textbf{BLEU}} & \textit{\textbf{\#Dial.}} & \textit{\textbf{\#Utt.}} & \textit{\textbf{Edit}} & \textit{\textbf{BLEU}} \\ \hline
\textit{\textbf{Fr}} & 248 & 3868 & 21.23 & 94.45 & 249 & 3900 & 24.29 & 94.19 \\
\textit{\textbf{It}} & 140 & 2160 & 83.01 & 80.45 & 140 & 2192 & 81.93 & 80.08 \\
\textit{\textbf{Id}} & 484 & 7562 & 157.58 & 60.46 & 484 & 7540 & 156.19 & 60.66 \\
\textit{\textbf{Jp}} & 275 & 4278 & 71.41 & 53.66 & 275 & 4322 & 75.83 & 49.56 \\
\textit{\textbf{Ko}} & 299 & 4684 & 74.04 & 61.25 & 300 & 4678 & 70.96 & 62.49 \\
\textit{\textbf{Zh}} & 222 & 3440 & 30.33 & 59.89 & 222 & 3458 & 33.07 & 64.61 \\ \hline
\end{tabular}
}
\caption{The statistics of the collected dataset. We report the number of dialogues (\#Dial.) and utterances (\#Utt.) of the validation and test set in six languages. Edit distance per dialogue (Edit) and BLEU score are computed to show the difference between the human-annotated dataset and auto-translated dataset. (Training set is reported in Appendix A)}
\label{table:stat}
\end{table*}

\section{Related Work}

\paragraph{Dialogue Systems} are categorized as goal-oriented~\citep{williams2007partially,young2013pomdp} and chit-chat~\citep{serban2016generative,vinyals2015neural}. Interested readers may refer to \citet{gao2018neural} for a general overview. In this paper, we focus on the latter, for which,  in recent years, several tasks and datasets have been proposed to ground the conversation on knowledge~\cite{dinan2018wizard,gopalakrishnan2019topical,shuster2018engaging,fan2019eli5,reddy2019coqa,choi2018quac,moon2019opendialkg} such as Wiki-Articles, Reddit-Post, and CNN-Article. In this work, we focus on personalized dialogue agents where the dialogues are grounded on persona information.

\citet{li2016persona} was the first to introduce a persona-grounded dialogue dataset for improving response consistency. Later on, \citet{personachat} and \citet{dinan2019second} introduced Persona-chat, a multi-turn conversational dataset, where two speakers are paired, and a persona description (4--5 sentences) is randomly assigned to each of them. By conditioning the response generation on the persona descriptions, a chit-chat model is able to produce a more persona-consistent dialogue~\cite{personachat}. Several works have improved on the initial baselines with various methodologies~\cite{kulikov2018importance,yavuz2019deepcopy,hancock2019learning,madotto2019personalizing,joshi2017personalization,zemlyanskiy2018aiming}, especially using large pre-trained models~\cite{wolf2019transfertransfo,zhang2019dialogpt}.

\paragraph{Multilingual}
Extensive approaches have been introduced to construct multilingual systems, for example, multilingual semantic role labeling~ \cite{akbik2015generating,he2019syntax}, multilingual machine translation~\cite{johnson2017google}, multilingual automatic speech recognition~\cite{toshniwal2018multilingual,yue2019end,nakayama2019zero,winata2019code}, and named entity recognition~\cite{winata2019learning, winata2019hierarchical}. 
Multilingual deep contextualized model such as Multilingual BERT (M-BERT)~\cite{devlin2018bert} have been commonly used to represent multiple languages and elevate the performance in many NLP applications, such as classification tasks~\cite{pires2019multilingual}, textual entailment, named entity recognition~\cite{K2020CrossLingual}, and natural language understanding~\cite{liu2019attentioninformed}. Multilingual datasets have also been created for a number of NLP tasks, such as named entity recognition or linking~\cite{sang2002introduction,sang2003introduction,pan2017cross,aguilar2018named}, question answering~\cite{liu2019xqa,lewis2019mlqa}, semantic role labeling~\cite{hajic2009conll}, part-of-speech tagging~\cite{nivre2017universal}, dialogue state tracking~\cite{mrkvsic2017semantic}, and natural language understanding~\cite{schuster2019cross}. 
However, none of these datasets include the multilingual chit-chat task.
% On another line of work, \citet{mrkvsic2017semantic} expanded the monolingual dialogue state tracking annotation of the WOZ 2.0 dataset~\cite{mrkvsic2017neural} into two additional languages, namely German and Italian. 

 \begin{figure*}[t]
    \centering
        \input{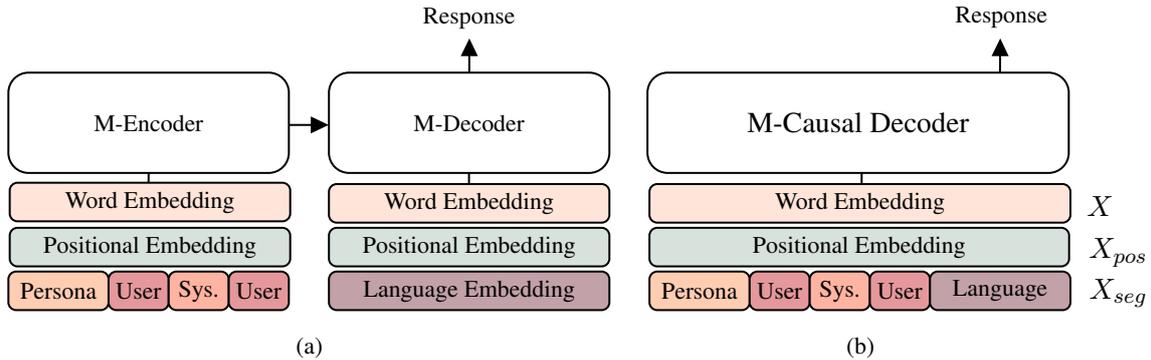}
    \caption{(a) Multilingual Encoder-Decoder model. (b) Multilingual Causal Decoder model. (Detailed illustration is reported in Appendix B)}
    \label{fig:MBERT}
\end{figure*}

\paragraph{Cross-lingual}
% Add stuff about multilingual in \textbf{DIALOGUE RESEARCH HERE}, for example, multilingual NLU and DST here.
Cross-lingual adaptation learns the inter-connections among languages and circumvents the requirement of extensive training data in target languages~\cite{wisniewski2014cross,zhang2016ten,liu2019zero}. Cross-lingual transfer learning methods have been applied to multiple NLP tasks, such as named entity recognition~\cite{ni2017weakly,xie2018neural}, natural language understanding ~\cite{liu2019attentioninformed}, dialogue state tracking~\cite{chen2018xl}, part-of-speech tagging~\cite{wisniewski2014cross,zhang2016ten,kim2017cross}, and dependency parsing~\cite{ahmad2019difficulties,schuster2019crosslingual}. Meanwhile, \citet{lample2019cross} and \citet{conneau2019unsupervised} proposed pre-trained cross-lingual language models to align multiple language representations, achieving state-of-the-art results in many cross-lingual classification tasks. The aforementioned tasks focused on classification and sequence labeling, while instead, \citet{chi2019cross} proposed to pre-train both the encoder and decoder of a sequence-to-sequence model (XNLG) to conduct cross-lingual generation tasks, namely, question generation and abstractive summarization. The latter is the closest to our task since it focuses on language generation; however cross-lingual dialogue generation has not yet been explored.

\section{Data Collection}
\label{data}
The proposed XPersona dataset is an extension of the persona-chat dataset~\cite{personachat,dinan2019second}. Specifically, we extend the ConvAI2~\cite{dinan2019second} to six languages: Chinese, French, Indonesian, Italian, Korean, and Japanese. Since the test set of ConvAI2 is hidden, we split the original validation set into a new validation set and test sets. Then, we firstly automatically translate the training, validation, and test set using APIs (PapaGo~\footnote{https://papago.naver.com} for Korean, Google Translate~\footnote{https://translate.google.com} for other languages). For each language, we hired native speaker annotators with a fluent level of English and asked them to revise the machine-translated dialogues and persona sentences in the validation set and test set according to original English dialogues. The main goal of human annotation is to ensure the revised conversations are coherent and fluent in target language despite the cultural discrepancy in different languages. Therefore, annotators are not restricted to translate the English dialogues. They are also \textbf{allowed} to customize dialogues and persona sentences. The annotated dialogues can deviate from original translation while \textbf{retain persona and conversation consistency}. The full annotation instructions are reported in Appendix A.

Compared to collecting new persona sentences and dialogues in each language, human-annotating the dialogues by leveraging translation APIs has multiple advantages. First, it increases the data distribution similarity across languages~\cite{conneau2018xnli}, which can better examine the system's cross-lingual transferability. Second, revising the machine-translated dialogues based on the original English dialogue improves the data construction efficiency. Third, it leverages the well-constructed English persona conversations as a reference to ensure the dialogue quality without the need for training a new pool of workers to generate new samples~\cite{conneau2018xnli}. 

On the other hand, human-translating the entire training-set ($\sim$130K utterances) in six languages is expensive. Therefore, we propose an iterative method to improve the quality of the automatically translated training set. We firstly sample 200 dialogues from the training set ($\sim$2600 utterances) in each language, and we assign human annotators to list all frequent translation mistakes in the given dialogues. For example, daily colloquial English expressions such as ``cool", ``I see", and ``lol" are usually literally translated. After that, we use a simple string matching to revise the inappropriate translations in the whole training-set and return a revision log, which records all the revised utterances. Then, we assign human annotators to check all the revised utterances and list translation mistakes again. We repeat this process at least twice for each language. Finally, we summarize the statistics of the collected dataset in Table \ref{table:stat}.

\section{Multilingual Personalized Conversational Models}
% \subsection{Notation}
Let us define a dialogue $\mathcal{D}=\{U_1,S_1,U_2,S_2, \dots, U_n, S_n\}$ as an alternating set of utterances from two speakers, where $U$ and $S$ represent the user and the system, respectively. Each speaker has its corresponding persona description that consists of a set of sentences $\mathcal{P}=\{P_1,\dots,P_m\}$. Given the system persona sentences $\mathcal{P}_s$ and dialogue history $\mathcal{D}_t=\{U_1,S_1,U_2, \dots,S_{t-1}, U_t\}$, we are interested in predicting the system utterances $S_t$.

\subsection{Model Architecture}
\label{multi_section}
We explore both encoder-decoder and causal decoder architectures, and we leverage existing pre-trained contextualized multilingual language models as weights initialization. Hence, we firstly define the multilingual embedding layer and then the two multilingual models used in our experiments.    
% and explore the encoder-decoder (M-EncDec) and decoder only (M-DecOnly) model architectures.

\paragraph{Embedding}
We define three embedding matrices: word embedding $E^W\in \mathbb{R}^{|V| \times d}$, positional embedding $E^P\in \mathbb{R}^{M \times d}$, and segmentation embedding $E^S\in \mathbb{R}^{|S| \times d}$, where $|.|$ denotes set cardinality, $d$ is the embedding size, $V$ denotes the vocabulary, $M$ denotes the maximum sequence length, and $S$ denotes the set of segmentation tokens. Segmentation embedding~\cite{wolf2019transfertransfo} is used to indicate whether the current token is part of i) \textbf{Persona} sentences, ii) System (\textbf{Sys.}) utterances, iii) \textbf{User} utterances, iv) response in \textbf{Language} $l_{id}$.
% The segmentation tokens $S$ includes a special token for Persona, System (Sys.), and User sentences for modeling multi-turn dialogue structure~\cite{wolf2019transfertransfo}, and a language id ($l_{id}$) for each of the language. 
The language embedding $l_{id}$ is used to inform the model which language to generate. Hence, given a sequence of tokens $X$, the embedding functions $E$ are defined as:
\begin{equation}
    E(X) = E^W(X) \oplus E^P(X_{pos}) \oplus E^S(X_{seg}),
\end{equation}
where $\oplus$ denotes the positional sum, $X_{pos}=\{1,\dots,|X|\}$ and $X_{seg}$ is the sequence of segmentation tokens, as in \citet{wolf2019transfertransfo}. Figure~\ref{fig:MBERT} shows a visual representation of the embedding process. A more detailed illustration is reported in Appendix B. 

\paragraph{Encoder-Decoder} To model the response generation, we use a Transformer~\cite{vaswani2017attention} based encoder-decoder~\cite{vinyals2015neural}. 
As illustrated in Figure \ref{fig:MBERT}, we concatenate~\footnote{We use the notation $[a; b]$ for concatenating the vectors $a$ and $b$} the system persona $\mathcal{P}_s$ with the dialogue history $\mathcal{D}_t$. Then we use the embedding layer $E$ to finally pass it to the encoder. In short, we have:
\begin{equation}
    H = \text{Encoder}(E([\mathcal{P}_s;\mathcal{D}_t])),
\end{equation} 
where $H \in \mathbb{R}^{L \times d_{model}}$ is the hidden representation computed by the encoder, and $L$ denotes the input sequence length. Then, the decoder attends to $H$ and generates the system response $S_t$ token by token. In the decoder, segmentation embedding is the language ID embedding (e.g., we look up the embedding for Italian to decode Italian). Thus:
\begin{equation}
    S_t = \text{Decoder}_t(H,l_{id}),
\end{equation}

\begin{table*}[t]
\begin{tabular}{r|cc|cc|cc|cc|cc}
\hline
\multicolumn{1}{c|}{\multirow{2}{*}{}} & \multicolumn{2}{c|}{\textbf{Bert2Bert}} & \multicolumn{2}{c|}{\textbf{M-Bert2Bert}} & \multicolumn{2}{c|}{\textbf{CausalBert}} & \multicolumn{2}{c|}{\textbf{M-CausalBert}} & \multicolumn{2}{c}{\textbf{XNLG}} \\ \cline{2-11}
\multicolumn{1}{c|}{} & \textit{\textbf{ppl.}} & \textit{\textbf{BLEU}} & \textit{\textbf{ppl.}} & \textit{\textbf{BLEU}} & \textit{\textbf{ppl.}} & \textit{\textbf{BLEU}} & \textit{\textbf{ppl.}} & \textit{\textbf{BLEU}} & \textit{\textbf{ppl.}} & \textit{\textbf{BLEU}} \\ \hline
\textit{\textbf{En}} & 21.99          & 1.53         & 25.99          & 0.57          & 16.08           & 1.79           & \textbf{15.62}   & 1.97  & 54.74*         & \textbf{2.25*}        \\ 
\textit{\textbf{Zh}} & 21.35          & 3.36         & 13.24          & 1.25          & \textbf{8.69}   & 5.51           & 9.27             & \textbf{5.7}  & 3482.27      & 2.16      \\ 
\textit{\textbf{It}} & 50.36          & 0.6          & 24.16          & 0.31          & 18.41           & 1.32            & \textbf{15.12}   & \textbf{1.3}  & 917.63            & 0.41         \\ 
\textit{\textbf{Jp}} & 10.09          & 5.23         & 10.64           & 0.79          & 11.00           & \textbf{6.74}  & \textbf{7.13}    & 4.53           & 999.81            & 0.0         \\ 
\textit{\textbf{Ko}} & 12.81          & 0.24            & 34.31          & 0.00          & 9.66            & 1.06           & \textbf{9.56}    & \textbf{1.08}  & -            & -         \\ 
\textit{\textbf{Id}} & 21.37          & 0.11         & 22.83          & 0.22          & 14.77           & \textbf{2.1}  & \textbf{14.61}   & 1.92           & 844.98            & 0.15         \\ 
\textit{\textbf{Fr}} & 13.22          & 0.35         & 15.58          & 0.50          & \textbf{10.39}   & 1.97           & 10.59              & \textbf{2.17}  & 640.33        & 0.09      \\ \hline
\end{tabular}
\caption{Results of automatic evaluation score on test set in seven languages. We compute the BLEU score and perplexity (ppl.) for monolingual, multilingual, and cross-lingual models.}
\label{table:automatic}
\end{table*}

\paragraph{Causal Decoder} As an alternative to encoder-decoders, the causal-decoders~\cite{radford2018improving,radford2019language,he2018layer} have been used to model conversational responses~\cite{wolf2019transfertransfo,zhang2019dialogpt} by giving as a prefix the dialogue history. In our model, we concatenate the persona $\mathcal{P}_s$ and the dialogue history $\mathcal{D}_t$ as the language model prefix, and autoregressively decode the system response $S_t$ based on language embedding (i.e. $l_{id}$):
\begin{equation}
    S_t = \text{Decoder}(E([\mathcal{P}_s;\mathcal{D}_t]),l_{id}).
\end{equation}

Figure~\ref{fig:MBERT} shows the conceptual differences between the encoder-decoder and casual decoder. Note that in both multilingual models, the dialogue history encoding process is language-agnostic, while decoding language is controlled by the language embedding. Such design allows the model to understand mixed-language dialogue contexts and to responds in the desired language (details in Section \ref{discussion}).

\subsection{Training Strategy}
We consider two training strategies to learn a multilingual conversational model: multilingual training and cross-lingual training.

\paragraph{Multilingual Training} jointly learns to perform personalized conversations in multiple languages. We follow a transfer learning approach~\cite{wolf2019transfertransfo,see2019makes} by initializing our models with the weights of the large multilingual pretrained model M-Bert~\cite{pires2019multilingual}. For the causal decoder, we add the causal mask into self-attention layer to convert M-Bert encoder to decoder. For encoder-decoder model, we randomly initialize the cross encoder-decoder attention~\cite{rothe2019leveraging}. Then, we train the both models on the combined training set in all 7 languages using cross-entropy loss.

\paragraph{Cross-lingual Training} transfers knowledge from the source language data to the target languages. In this setting, the model is trained on English (source language) conversational samples, and evaluated on the other 6 languages. Following the methodology proposed by ~\citet{chi2019cross}, we align the embedded representations of different languages into the same embedding space by applying cross-lingual pre-training to the encoder-decoder model. The pre-training procedure consists of two stages:
\begin{itemize}
    \item pre-training the encoder and the decoder independently utilizing masked language modeling, as in ~\citet{lample2019cross};
    \item jointly pre-training the encoder-decoder by using two objective functions: Cross-Lingual Auto-Encoding (XAE) and Denoising Auto-Encoding (DAE)~\cite{chi2019cross}.
\end{itemize}
For instance, DAE adds perturbations to the input sentence of encoder and tries to reconstructs the original sentence using the decoder, whereas, XAE uses parallel translation data to pre-train both the encoder and decoder with machine translation objective. As in the multilingual models, the language IDs are fed into the decoder to control the language of generated sentences. Both pre-training stages require both parallel and non-parallel data in the target language.

After the two stages of pre-training, the model is fine-tuned using just the source language samples (i.e., English) with the same cross-entropy loss as for the multilingual training. However, as suggested in~\citet{chi2019cross}, only the encoder parameters are updated with back-propagation and both the decoder and the word embedding layer remain frozen. This retains the decoders' ability to generate multilingual output while still being able to learn new tasks using only the target language.

\begin{table*}[t]
\begin{tabular}{c|c|ccc|ccc|ccc}
\hline
\multirow{9}{*}{\rotatebox{90}{\textbf{Multi Wins \%}}} & \multirow{2}{*}{\textit{\textbf{Lang}}} & \multicolumn{3}{c|}{\textit{\textbf{Engageness}}}      & \multicolumn{3}{c|}{\textit{\textbf{Interestingness}}} & \multicolumn{3}{c}{\textit{\textbf{Humanness}}}       \\ \cline{3-11} 
        &                                             & \textbf{Human}   & \textbf{Mono}    & \textbf{Poly}    & \textbf{Human}   & \textbf{Mono}    & \textbf{Poly}    & \textbf{Human}   & \textbf{Mono}    & \textbf{Poly}    \\ \cline{2-11} 
        & \textit{\textbf{En}}                        & \textbf{23.33} & \textbf{68.57} & 36.36          & \textbf{23.33} & \textbf{64.29} & \textbf{32.73} & \textbf{30.00} & \textbf{62.86} & 42.73          \\ 
        & \textit{\textbf{Fr}}                        & 32.00                 & 55.17          & 42.86          & \textbf{16.00}        & 53.45          & 48.21          & \textbf{28.00}        & 50.00          & 44.64          \\ 
        & \textit{\textbf{Id}}                        & \textbf{21.67} & 51.67          & \textbf{65.45} & \textbf{23.33} & 46.67          & 55.45          & \textbf{25.00} & 46.67          & \textbf{65.45} \\ 
        & \textit{\textbf{It}}                        & \textbf{35.00} & 48.33          & 56.36          & \textbf{30.00} & 48.33          & 53.64          & \textbf{30.00} & 40.00          & 57.27          \\ 
        & \textit{\textbf{Jp}}                        & \textbf{18.33} & 50.00          & \textbf{61.82} & \textbf{13.33} & 43.33          & 45.45          & \textbf{18.33} & 51.67          & 59.09          \\ 
        & \textit{\textbf{Ko}}                        & \textbf{30.00} & 52.46          & \textbf{62.39} & \textbf{26.67} & 50.82          & 59.63          & \textbf{28.33} & 52.46          & \textbf{64.22} \\ 
        & \textit{\textbf{Zh}}                        & \textbf{36.67} & 55.00          & \textbf{65.45} & \textbf{36.67} & 60.00          & \textbf{61.82} & \textbf{36.67} & 55.00          & \textbf{70.91} \\ \hline
\end{tabular}
\caption{Results of ACUTE-EVAL human evaluation. Tests are conducted pairwise between M-CausalBert (Multi.) and other models (Human, Poly-encoder (Poly), Monolingual CausalBert (Mono)). Numbers indicate the winning rate of Multi. Numbers in bold are statistically significant ($p<0.05$).}
\label{table:human}
\end{table*}

\section{Experiments}

\subsection{Evaluation Metrics}
Evaluating open-domain chit-chat models is challenging, especially in multiple languages and at the dialogue-level. Hence, we evaluate our models using both automatic and human evaluation. In both cases, human-annotated dialogues are used, which show the importance of the provided dataset.

\paragraph{Automatic}
For each language, we evaluate responses generated by the models using perplexity (ppl.) and BLEU~\cite{papineni2002bleu} with reference to the human-annotated responses.  Although these automatic measures are not perfect~\cite{D16-1230}, they help to roughly estimate the performance of different models under the same test set. More recently, \citet{adiwardana2020towards} has shown the correlation between perplexity and human judgment in open-domain chit-chat models.

\paragraph{Human}
Asking humans to evaluate the quality of a dialogue model is challenging, especially when multiple models have to be compared. The likert score (a.k.a. 1 to 5 scoring) has been widely used to evaluate the interactive experience with conversational models~\cite{venkatesh2018evaluating,see2019makes,personachat,dinan2019second}. In such evaluation, a human interacts with the systems for several turns, and then they assign a score from 1 to 5 based on three questions~\cite{personachat} about fluency, engagingness, and consistency. This evaluation is both expensive to conduct and requires many samples to achieve statistically significant results~\citet{li2019acute}. To cope with these issues, \citet{li2019acute} proposed ACUTE-EVAL, an A/B test evaluation for dialogue systems. The authors proposed two modes: human-model chats and self-chat~\cite{li2016deep,ghandeharioun2019approximating}. In this work, we opt for the latter since it is cheaper to conduct and achieves similar results~\cite{li2019acute} to the former. Another advantage of using this method is the ability to evaluate multi-turn conversations instead of single-turn responses.

Following ACUTE-EVAL, the annotator is provided with two full dialogues made by self-chat or human-dialogue. The annotator is asked to choose which of the two dialogues is better in terms of engagingness, interestingness, and humanness. For each comparison, we sample 60--100 conversations from both models. In Appendix C, we report the exact questions and instructions given to the annotators, and the user interface used in the evaluation. We hired native speakers annotators for all six considered languages. The annotators were different from the dataset collection annotators to avoid any possible bias.

\subsection{Implementation Details}
\paragraph{Multilingual Models} We use the "BERT-Base, Multilingual Cased" checkpoint, and we denote the multilingual encoder-decoder model as \textbf{M-Bert2Bert} ($\sim$220M parameters) and causal decoder model as \textbf{M-CausalBert} ($\sim$110M parameters). We fine-tune both models in the combined training set (English in Persona-chat~\cite{personachat}, six languages in Xpersona) for five epochs with AdamW~\footnote{AdamW: Adam algorithm with weight decay} optimizer and a learning rate of $6.25e$-$5$.

\paragraph{Monolingual Models} To verify whether the multilingual agent will under-perform the monolingual agent in the monolingual conversational task, we build a monolingual encoder-decoder model and causal decoder model for each language. For a fair comparison, we initialize the monolingual models with a pre-trained monolingual BERT~\footnote{The monolingual BERT pre-trained models are available in https://github.com/huggingface/transformers }~\cite{devlin2018bert,cui2019pre,martin2019camembert}. We denote the monolingual encoder-decoder model as \textbf{Bert2Bert} ($\sim$220M parameters) and causal decoder model as \textbf{CausalBert} ($\sim$110M parameters). Then we fine-tune each model in each language independently for the same number of epoch and optimizer as the multilingual model. 

\paragraph{Translation-based Models} Another strong baseline we compare with is Poly-encoder~\cite{humeau2019poly}, a large-scale pre-trained retrieval model that has shown state-of-the-art performance in the English Persona-chat dataset~\cite{li2019acute}. We adapt this model to the other languages by using the Google Translate API to translate target languages (e.g., Chinese) query to English as the input to the model, then translate the English response back to the target language. Thus, the response generation flow is: target query $\rightarrow$ English query $\rightarrow$ English response $\rightarrow$ target response. We denote this model as \textbf{Poly}.

\paragraph{Cross-lingual Models.} 
In the first pre-training stage, we use the pre-trained weights from XLMR-base~\cite{conneau2019unsupervised}. Then, we follow the second pre-training stage of XNLG~\cite{chi2019cross} for pre-training Italian, Japanese, Korean, Indonesia cross-lingual transferable models. For Chinese and French, we directly apply the pre-trained XNLG~\cite{chi2019cross} weights\footnote{Available in https://github.com/CZWin32768/XNLG}. Then, the pre-trained models are fine-tune on English PersonaChat training set and early stop based on the perplexity on target language validation set.

% We follow the pre-trained stages from XNLG~\cite{chi2019cross}, and utilize the pre-trained weight to initialize the encoder-decoder model.
% Then, we fine-tune the model on English training samples and use an early stop strategy based on the perplexity on the translated target language development set to select the model. Finally, we conduct zero-shot adaptation to target languages.
% \citet{chi2019cross} use XLM~\cite{lample2019cross} for the first stage of the cross-lingual pre-training and only Chinese and French in our dataset are included in the XLM model. Hence, we utilize the available pre-trained weight\footnote{Available in https://github.com/CZWin32768/XNLG} 
% % that contains ten encoder layers and six decoder layers
% for fine-tuning the model to Chinese and French.
% And for the other languages, we leverage the pre-trained XLM-RoBERTa~\cite{conneau2019unsupervised} for the first stage and follow the same method for the second stage pre-training.

% \begin{figure*}
%     \centering
% {\footnotesize
% \begin{minipage}[b]{0.32\textwidth}
% \includegraphics[width=\textwidth]{En.pdf}
% \end{minipage}
% \vline
% \begin{minipage}[b]{0.32\textwidth}
% \includegraphics[width=\textwidth]{Zh.pdf}
% \end{minipage}
% \vline
% \begin{minipage}[b]{0.32\textwidth}
% \includegraphics[width=\textwidth]{IT.pdf}
% \end{minipage}
% }
%     \caption{Caption}
%     \label{fig:my_label}
% \end{figure*}

\subsection{Results and Discussion}
\subsubsection{Quantitative Analysis}
Table~\ref{table:automatic} compares monolingual, multilingual, and cross-lingual models in terms of BLEU and perplexity in the human-translated test set. On both evaluation matrices, the causal decoder models outperform the encoder-decoder models. We observe that the encoder-decoder model tends to overlook dialogue context and generate digressive responses. (Generated samples are available in Appendix D) We hypothesize that this is because the one-to-many problem~\cite{zhao2017learning} in open-domain conversation weakens the relation between encoder and decoder; thus the well pre-trained decoder (Bert) easily converges to a locally-optimal, and learns to ignore the dialogue context from the encoder and generate the response in an unconditional language model way. We leave the investigation of this problem to future work.  On the other hand, M-CausalBert achieves a comparable or slightly better performance compared to CausalBert, which suggests that M-CausalBert leverages the data from other languages. As expected, we observe a significant gap between the cross-lingual model and other models, which indicates that cross-lingual zero-shot conversation modeling is very challenging. 

Table~\ref{table:human} shows the human evaluation result of comparing M-CausalBert (Multi) against the human, translation-based Poly-encoder (Poly), and monolingual CausalBert (Mono). The results illustrate that Multi outperforms Mono in English and Chinese, and is on par with Mono in other languages. On the other hand, Poly shows a strong performance in English as it was pre-trained with a large-scale English conversation corpus. In contrast, the performance of Poly drops in other languages, which indicates that the imperfect translation affects translation-based systems. We also conduct M-CausalBert (Multi) against XNLG (cross) human evaluation, and Multi achieve nearly 100 percent winning rate.

\begin{table}[!t]
\resizebox{0.99\linewidth}{!}{
\begin{tabular}{ll}
\multicolumn{2}{c}{\textbf{System Persona}} \\
\multicolumn{2}{c}{\begin{tabular}[c]{@{}c@{}}My favorite activity is rock climbing. \\ I'm a mechanical engineer. \\ I like to go hiking in the spring. \\ I live in colorado.\end{tabular}} \\
\multicolumn{2}{c}{\textbf{Dialogue}} \\ \hline
% {\color[HTML]{CB0000} Usr} & ciao come stai? (hi, how are you?) \\ \hline
% {\color[HTML]{32CB00} Sys} & what are you up to this morning? \\ \hline
{\color[HTML]{CB0000} Usr} & \begin{tabular}[c]{@{}l@{}}studio inglese oggi e tu? (I am studying \\ English today, you?)\end{tabular} \\ \hline
{\color[HTML]{32CB00} Sys} & just watching netflix with friends \\ \hline
{\color[HTML]{CB0000} Usr} & \begin{tabular}[c]{@{}l@{}}\begin{CJK*}{UTF8}{gbsn}我喜欢攀岩，你呢？\end{CJK*} (I like rock \\ climbing, what about you?)\end{tabular} \\ \hline
{\color[HTML]{32CB00} Sys} & \begin{tabular}[c]{@{}l@{}}nice, I enjoy it while in the summer. \\ what are you up to?\end{tabular} \\ \hline
{\color[HTML]{CB0000} Usr} & \begin{tabular}[c]{@{}l@{}}saya ingin melakukan panjat tebing \\ (I want to go to hike)\end{tabular} \\ \hline
{\color[HTML]{32CB00} Sys} & I'm going hiking the most time of year \\ \hline
{\color[HTML]{CB0000} Usr} & \begin{tabular}[c]{@{}l@{}}\begin{CJK*}{UTF8}{min}スポーツがお好きなようですね \end{CJK*} \\ (It seems you like sports)\end{tabular} \\ \hline
{\color[HTML]{32CB00} Sys} & \begin{tabular}[c]{@{}l@{}}I love the outdoors too, do you have \\ any hobbies?\end{tabular} \\ \hline
{\color[HTML]{CB0000} Usr} & \begin{tabular}[c]{@{}l@{}}\begin{CJK*}{UTF8}{}\CJKfamily{mj}\text{나는 }  \text{티비  } \text{보는 } \text{것을 } \text{좋아해}.\end{CJK*}(I like \\ to watch tv)\end{tabular} \\ \hline
{\color[HTML]{32CB00} Sys} & i really like hiking and listening to music \\ \hline
\end{tabular}
}
\caption{Many-to-one: understand mixed-language dialogue context in multiple languages and generate response in one language}
\label{table:many2one}
\end{table}

\begin{table}[t]
\resizebox{0.99\linewidth}{!}{
\begin{tabular}{ll}
\multicolumn{2}{c}{\textbf{System Persona}} \\
\multicolumn{2}{c}{\begin{tabular}[c]{@{}c@{}}I love to drink fancy tea. \\ I have a big library at home. \\ I'm a museum tour guide. \\ I'm partly deaf.\end{tabular}} \\
\multicolumn{2}{c}{\textbf{Dialogue}} \\ \hline
{\color[HTML]{CB0000} Usr} & Hi, I am a computer science student, you? \\ \hline
{\color[HTML]{32CB00} En} & \begin{tabular}[c]{@{}l@{}}I'm a bookkeeper for the local \\ museum of art.\end{tabular} \\ \hline
{\color[HTML]{32CB00} Zh} & \begin{tabular}[c]{@{}l@{}}\begin{CJK*}{UTF8}{gbsn}你好，我是一名博物馆老师。\end{CJK*} \\ (I am a teacher in a museum)\end{tabular} \\ \hline
{\color[HTML]{32CB00} Fr} & \begin{tabular}[c]{@{}l@{}}bonjour, je suis juste un séjour à la \\ maison maman de mon immense \\ bibliothèque. (hello, I'm just a stay \\ at home my huge library.)\end{tabular} \\ \hline
{\color[HTML]{32CB00} It} & \begin{tabular}[c]{@{}l@{}}Sono un bibliotecario, ma ho bisogno \\ di rilassarmi. (I am a bookkeper, but \\ I need to relax)\end{tabular} \\ \hline
{\color[HTML]{32CB00} Jp} & \begin{tabular}[c]{@{}l@{}}\begin{CJK*}{UTF8}{min}こんにちは。私は大学にいます、\end{CJK*}\\  \begin{CJK*}{UTF8}{min}あなたの専攻は何ですか？ \end{CJK*} \\ (Hello. I am in college, what is your major?)\end{tabular} \\ \hline
{\color[HTML]{32CB00} Id} & \begin{tabular}[c]{@{}l@{}}Saya tidak tahu, tetapi saya tuli. \\ (I don't know I am deaf)\end{tabular} \\ \hline
{\color[HTML]{32CB00} Ko} & \begin{tabular}[c]{@{}l@{}}\begin{CJK*}{UTF8}{}\CJKfamily{mj}\text{안녕, } \text{나는 } \text{선생님이야.} \end{CJK*}\\ \begin{CJK*}{UTF8}{}\CJKfamily{mj}\text{너는 } \text{무엇을 } \text{공부하고 }\text{ 있니?}\end{CJK*}\\ (Hello, I am a teacher. What are you studying?)\end{tabular} \\ \hline
\end{tabular}
}

\caption{One-to-many: response one dialogue context with 7 different languages}
\label{table:one2many}
\end{table}

\subsubsection{Qualitative Analysis and Discussion}
\label{discussion}
We randomly sample 7 self-chat dialogues for each baseline model in the seven languages and report them in Appendix D., And we summarize the generation of each model as follows:
\paragraph{Poly} Poly-encoder, pretrained on 174 million Reddit data, can accurately retrieve coherent and diverse responses in English. However, in the other six languages, some of the retrieved responses are digressive due to translation error.

\paragraph{Monolingual \& Multilingual} We observe that both the monolingual and multilingual models can generate fluent responses. Compared to Bert2Bert and M-Bert2Bert, CausalBert and M-CausalBert can generate more on-topic responses but sometimes repeat through turns. CausalBert and M-CausalBert are on par with each other in monolingual conversational tasks, while M-CausalBert shows the advantage of handling a mixed-language context. For multilingual speakers, the conversation may involve multiple languages. Therefore, we experiment on M-CausalBert with two settings: 1) many-to-one, in which users converse with the model in 6 languages, and the model generate responses in English, 2) one-to-many, in which users converse with the model using English, and the model generates responses in 6 languages using language embedding and corresponding persona sentences. Table \ref{table:many2one} and table \ref{table:one2many} illustrate the generation examples under these settings (more examples reported in Appendix C.1). Most of the time, M-CausalBert can understand the mixed-language context, and decode coherent response in different languages. Understanding the mixed-language dialogue context is a desirable skill for end-to-end chit-chat systems, and a systematic study of this research question is needed in future.

\paragraph{Cross-lingual.} The current state-of-the-art cross-lingual generation approach XNLG~\cite{chi2019cross} shows inferior performance on multi-turn dialogue tasks, and generates repetitive responses. Although cross-lingual dialogue generation is challenging, it reduces the human effort for data annotation in different languages. Therefore, the cross-language transfer is an important direction to investigate.

\section{Conclusion}
% Multilingual dialogue systems is important research question.
In this paper, we studied both cross-lingual and multilingual approaches in end-to-end personalized dialogue modeling. We presented the XPersona dataset, a multilingual extension of Persona-Chat, for evaluating the multilingual personalized chatbots. We further provided both cross-lingual and multilingual baselines and compared them with the monolingual approach and two-stage translation approach. Extensive automatic evaluation and human evaluation were conducted to examine the models' performance.
The experimental results showed that multilingual trained models, with a single model across multiple languages, can outperform the two-stage translation approach and is on par with monolingual models. On the other hand, the current state-of-the-art cross-lingual approach XNLG achieved lower performance than other baselines. In future work, we plan to research a more advanced cross-lingual generation approach and construct a mixed-language conversational benchmark for evaluating multilingual systems.

%
% On the other hand, state-of-the-art cross-lingual trained models achieved the worst results, showing that cross-lingual conversation modeling is a very challenging task. We hope that our dataset and baselines will accelerate research in multilingual dialogue systems.
% that the performance of the cross-lingual approach is far from ideal performance, and further 

\bibliography{emnlp2020}

\begin{thebibliography}{77}
\expandafter\ifx\csname natexlab\endcsname\relax\def\natexlab#1{#1}\fi

\bibitem[{Adiwardana et~al.(2020)Adiwardana, Luong, So, Hall, Fiedel,
  Thoppilan, Yang, Kulshreshtha, Nemade, Lu et~al.}]{adiwardana2020towards}
Daniel Adiwardana, Minh-Thang Luong, David~R So, Jamie Hall, Noah Fiedel, Romal
  Thoppilan, Zi~Yang, Apoorv Kulshreshtha, Gaurav Nemade, Yifeng Lu, et~al.
  2020.
\newblock Towards a human-like open-domain chatbot.
\newblock \emph{arXiv preprint arXiv:2001.09977}.

\bibitem[{Aguilar et~al.(2018)Aguilar, AlGhamdi, Soto, Diab, Hirschberg, and
  Solorio}]{aguilar2018named}
Gustavo Aguilar, Fahad AlGhamdi, Victor Soto, Mona Diab, Julia Hirschberg, and
  Thamar Solorio. 2018.
\newblock Named entity recognition on code-switched data: Overview of the calcs
  2018 shared task.
\newblock In \emph{Proceedings of the Third Workshop on Computational
  Approaches to Linguistic Code-Switching}, pages 138--147.

\bibitem[{Ahmad et~al.(2019)Ahmad, Zhang, Ma, Hovy, Chang, and
  Peng}]{ahmad2019difficulties}
Wasi Ahmad, Zhisong Zhang, Xuezhe Ma, Eduard Hovy, Kai-Wei Chang, and Nanyun
  Peng. 2019.
\newblock On difficulties of cross-lingual transfer with order differences: A
  case study on dependency parsing.
\newblock In \emph{Proceedings of the 2019 Conference of the North American
  Chapter of the Association for Computational Linguistics: Human Language
  Technologies, Volume 1 (Long and Short Papers)}, pages 2440--2452.

\bibitem[{Akbik et~al.(2015)Akbik, Chiticariu, Danilevsky, Li, Vaithyanathan,
  and Zhu}]{akbik2015generating}
Alan Akbik, Laura Chiticariu, Marina Danilevsky, Yunyao Li, Shivakumar
  Vaithyanathan, and Huaiyu Zhu. 2015.
\newblock Generating high quality proposition banks for multilingual semantic
  role labeling.
\newblock In \emph{Proceedings of the 53rd Annual Meeting of the Association
  for Computational Linguistics and the 7th International Joint Conference on
  Natural Language Processing (Volume 1: Long Papers)}, pages 397--407.

\bibitem[{Chen et~al.(2018)Chen, Chen, Su, Wang, Yu, Yan, and
  Wang}]{chen2018xl}
Wenhu Chen, Jianshu Chen, Yu~Su, Xin Wang, Dong Yu, Xifeng Yan, and
  William~Yang Wang. 2018.
\newblock Xl-nbt: A cross-lingual neural belief tracking framework.
\newblock In \emph{Proceedings of the 2018 Conference on Empirical Methods in
  Natural Language Processing}, pages 414--424.

\bibitem[{Chi et~al.(2019)Chi, Dong, Wei, Wang, Mao, and Huang}]{chi2019cross}
Zewen Chi, Li~Dong, Furu Wei, Wenhui Wang, Xian-Ling Mao, and Heyan Huang.
  2019.
\newblock Cross-lingual natural language generation via pre-training.
\newblock \emph{arXiv preprint arXiv:1909.10481}.

\bibitem[{Choi et~al.(2018)Choi, He, Iyyer, Yatskar, Yih, Choi, Liang, and
  Zettlemoyer}]{choi2018quac}
Eunsol Choi, He~He, Mohit Iyyer, Mark Yatskar, Wen-tau Yih, Yejin Choi, Percy
  Liang, and Luke Zettlemoyer. 2018.
\newblock Quac: Question answering in context.
\newblock In \emph{Proceedings of the 2018 Conference on Empirical Methods in
  Natural Language Processing}, pages 2174--2184.

\bibitem[{Conneau et~al.(2019)Conneau, Khandelwal, Goyal, Chaudhary, Wenzek,
  Guzm{\'a}n, Grave, Ott, Zettlemoyer, and Stoyanov}]{conneau2019unsupervised}
Alexis Conneau, Kartikay Khandelwal, Naman Goyal, Vishrav Chaudhary, Guillaume
  Wenzek, Francisco Guzm{\'a}n, Edouard Grave, Myle Ott, Luke Zettlemoyer, and
  Veselin Stoyanov. 2019.
\newblock Unsupervised cross-lingual representation learning at scale.
\newblock \emph{arXiv preprint arXiv:1911.02116}.

\bibitem[{Conneau et~al.(2018)Conneau, Rinott, Lample, Williams, Bowman,
  Schwenk, and Stoyanov}]{conneau2018xnli}
Alexis Conneau, Ruty Rinott, Guillaume Lample, Adina Williams, Samuel Bowman,
  Holger Schwenk, and Veselin Stoyanov. 2018.
\newblock Xnli: Evaluating cross-lingual sentence representations.
\newblock In \emph{Proceedings of the 2018 Conference on Empirical Methods in
  Natural Language Processing}, pages 2475--2485.

\bibitem[{Cui et~al.(2019)Cui, Che, Liu, Qin, Yang, Wang, and Hu}]{cui2019pre}
Yiming Cui, Wanxiang Che, Ting Liu, Bing Qin, Ziqing Yang, Shijin Wang, and
  Guoping Hu. 2019.
\newblock Pre-training with whole word masking for chinese bert.
\newblock \emph{arXiv preprint arXiv:1906.08101}.

\bibitem[{Devlin et~al.(2018)Devlin, Chang, Lee, and
  Toutanova}]{devlin2018bert}
Jacob Devlin, Ming-Wei Chang, Kenton Lee, and Kristina Toutanova. 2018.
\newblock Bert: Pre-training of deep bidirectional transformers for language
  understanding.
\newblock \emph{arXiv preprint arXiv:1810.04805}.

\bibitem[{Dinan et~al.(2019{\natexlab{a}})Dinan, Logacheva, Malykh, Miller,
  Shuster, Urbanek, Kiela, Szlam, Serban, Lowe et~al.}]{dinan2019second}
Emily Dinan, Varvara Logacheva, Valentin Malykh, Alexander Miller, Kurt
  Shuster, Jack Urbanek, Douwe Kiela, Arthur Szlam, Iulian Serban, Ryan Lowe,
  et~al. 2019{\natexlab{a}}.
\newblock The second conversational intelligence challenge (convai2).
\newblock \emph{arXiv preprint arXiv:1902.00098}.

\bibitem[{Dinan et~al.(2019{\natexlab{b}})Dinan, Roller, Shuster, Fan, Auli,
  and Weston}]{dinan2018wizard}
Emily Dinan, Stephen Roller, Kurt Shuster, Angela Fan, Michael Auli, and Jason
  Weston. 2019{\natexlab{b}}.
\newblock \href {https://openreview.net/forum?id=r1l73iRqKm} {Wizard of
  wikipedia: Knowledge-powered conversational agents}.
\newblock In \emph{International Conference on Learning Representations}.

\bibitem[{Etherington(2019)}]{etherington_2019}
Darrell Etherington. 2019.
\newblock \href
  {https://techcrunch.com/2019/09/25/amazon-launches-multilingual-mode-for-using-alexa-in-multiple-languages-at-once/}
  {Amazon launches multilingual mode for using alexa in multiple languages at
  once}.

\bibitem[{Fan et~al.(2019)Fan, Jernite, Perez, Grangier, Weston, and
  Auli}]{fan2019eli5}
Angela Fan, Yacine Jernite, Ethan Perez, David Grangier, Jason Weston, and
  Michael Auli. 2019.
\newblock Eli5: Long form question answering.
\newblock \emph{arXiv preprint arXiv:1907.09190}.

\bibitem[{Gao et~al.(2018)Gao, Galley, and Li}]{gao2018neural}
Jianfeng Gao, Michel Galley, and Lihong Li. 2018.
\newblock Neural approaches to conversational ai.
\newblock In \emph{The 41st International ACM SIGIR Conference on Research \&
  Development in Information Retrieval}, pages 1371--1374. ACM.

\bibitem[{Ghandeharioun et~al.(2019)Ghandeharioun, Shen, Jaques, Ferguson,
  Jones, Lapedriza, and Picard}]{ghandeharioun2019approximating}
Asma Ghandeharioun, Judy~Hanwen Shen, Natasha Jaques, Craig Ferguson, Noah
  Jones, Agata Lapedriza, and Rosalind Picard. 2019.
\newblock Approximating interactive human evaluation with self-play for
  open-domain dialog systems.
\newblock In \emph{Advances in Neural Information Processing Systems}, pages
  13658--13669.

\bibitem[{Gopalakrishnan et~al.(2019)Gopalakrishnan, Hedayatnia, Chen,
  Gottardi, Kwatra, Venkatesh, Gabriel, Hakkani-T{\"u}r, and
  AI}]{gopalakrishnan2019topical}
Karthik Gopalakrishnan, Behnam Hedayatnia, Qinlang Chen, Anna Gottardi, Sanjeev
  Kwatra, Anu Venkatesh, Raefer Gabriel, Dilek Hakkani-T{\"u}r, and
  Amazon~Alexa AI. 2019.
\newblock Topical-chat: Towards knowledge-grounded open-domain conversations.
\newblock \emph{Proc. Interspeech 2019}, pages 1891--1895.

\bibitem[{Hajic et~al.(2009)Hajic, Ciaramita, Johansson, Kawahara, Mart{\'\i},
  M{\`a}rquez, Meyers, Nivre, Pad{\'o}, {\v{S}}t{\v{e}}p{\'a}nek
  et~al.}]{hajic2009conll}
Jan Hajic, Massimiliano Ciaramita, Richard Johansson, Daisuke Kawahara,
  M~Ant{\`o}nia Mart{\'\i}, Llu{\'\i}s M{\`a}rquez, Adam Meyers, Joakim Nivre,
  Sebastian Pad{\'o}, Jan {\v{S}}t{\v{e}}p{\'a}nek, et~al. 2009.
\newblock The conll-2009 shared task: Syntactic and semantic dependencies in
  multiple languages.
\newblock In \emph{Proceedings of the Thirteenth Conference on Computational
  Natural Language Learning (CoNLL 2009): Shared Task}, pages 1--18.

\bibitem[{Hancock et~al.(2019)Hancock, Bordes, Mazare, and
  Weston}]{hancock2019learning}
Braden Hancock, Antoine Bordes, Pierre-Emmanuel Mazare, and Jason Weston. 2019.
\newblock Learning from dialogue after deployment: Feed yourself, chatbot!
\newblock \emph{arXiv preprint arXiv:1901.05415}.

\bibitem[{He et~al.(2019)He, Li, and Zhao}]{he2019syntax}
Shexia He, Zuchao Li, and Hai Zhao. 2019.
\newblock Syntax-aware multilingual semantic role labeling.
\newblock In \emph{Proceedings of the 2019 Conference on Empirical Methods in
  Natural Language Processing and the 9th International Joint Conference on
  Natural Language Processing (EMNLP-IJCNLP)}, pages 5353--5362.

\bibitem[{He et~al.(2018)He, Tan, Xia, He, Qin, Chen, and Liu}]{he2018layer}
Tianyu He, Xu~Tan, Yingce Xia, Di~He, Tao Qin, Zhibo Chen, and Tie-Yan Liu.
  2018.
\newblock Layer-wise coordination between encoder and decoder for neural
  machine translation.
\newblock In \emph{Advances in Neural Information Processing Systems}, pages
  7944--7954.

\bibitem[{Humeau et~al.(2019)Humeau, Shuster, Lachaux, and
  Weston}]{humeau2019poly}
Samuel Humeau, Kurt Shuster, Marie-Anne Lachaux, and Jason Weston. 2019.
\newblock Poly-encoders: Transformer architectures and pre-training strategies
  for fast and accurate multi-sentence scoring.
\newblock \emph{CoRR abs/1905.01969. External Links: Link Cited by}, 2:2--2.

\bibitem[{Johnson et~al.(2017)Johnson, Schuster, Le, Krikun, Wu, Chen, Thorat,
  Vi{\'e}gas, Wattenberg, Corrado et~al.}]{johnson2017google}
Melvin Johnson, Mike Schuster, Quoc Le, Maxim Krikun, Yonghui Wu, Zhifeng Chen,
  Nikhil Thorat, Fernanda Vi{\'e}gas, Martin Wattenberg, Greg Corrado, et~al.
  2017.
\newblock Google’s multilingual neural machine translation system: Enabling
  zero-shot translation.
\newblock \emph{Transactions of the Association for Computational Linguistics},
  5:339--351.

\bibitem[{Joshi et~al.(2017)Joshi, Mi, and Faltings}]{joshi2017personalization}
Chaitanya~K Joshi, Fei Mi, and Boi Faltings. 2017.
\newblock Personalization in goal-oriented dialog.
\newblock \emph{arXiv preprint arXiv:1706.07503}.

\bibitem[{K et~al.(2020)K, Wang, Mayhew, and Roth}]{K2020CrossLingual}
Karthikeyan K, Zihan Wang, Stephen Mayhew, and Dan Roth. 2020.
\newblock \href {https://openreview.net/forum?id=HJeT3yrtDr} {Cross-lingual
  ability of multilingual bert: An empirical study}.
\newblock In \emph{International Conference on Learning Representations}.

\bibitem[{Kim et~al.(2017)Kim, Kim, Sarikaya, and
  Fosler-Lussier}]{kim2017cross}
Joo-Kyung Kim, Young-Bum Kim, Ruhi Sarikaya, and Eric Fosler-Lussier. 2017.
\newblock Cross-lingual transfer learning for pos tagging without cross-lingual
  resources.
\newblock In \emph{Proceedings of the 2017 Conference on Empirical Methods in
  Natural Language Processing}, pages 2832--2838.

\bibitem[{Kulikov et~al.(2018)Kulikov, Miller, Cho, and
  Weston}]{kulikov2018importance}
Ilya Kulikov, Alexander~H Miller, Kyunghyun Cho, and Jason Weston. 2018.
\newblock Importance of a search strategy in neural dialogue modelling.
\newblock \emph{arXiv preprint arXiv:1811.00907}.

\bibitem[{Lample and Conneau(2019)}]{lample2019cross}
Guillaume Lample and Alexis Conneau. 2019.
\newblock Cross-lingual language model pretraining.
\newblock \emph{arXiv preprint arXiv:1901.07291}.

\bibitem[{Lewis et~al.(2019)Lewis, O{\u{g}}uz, Rinott, Riedel, and
  Schwenk}]{lewis2019mlqa}
Patrick Lewis, Barlas O{\u{g}}uz, Ruty Rinott, Sebastian Riedel, and Holger
  Schwenk. 2019.
\newblock Mlqa: Evaluating cross-lingual extractive question answering.
\newblock \emph{arXiv preprint arXiv:1910.07475}.

\bibitem[{Li et~al.(2016{\natexlab{a}})Li, Galley, Brockett, Spithourakis, Gao,
  and Dolan}]{li2016persona}
Jiwei Li, Michel Galley, Chris Brockett, Georgios Spithourakis, Jianfeng Gao,
  and Bill Dolan. 2016{\natexlab{a}}.
\newblock A persona-based neural conversation model.
\newblock In \emph{Proceedings of the 54th Annual Meeting of the Association
  for Computational Linguistics (Volume 1: Long Papers)}, volume~1, pages
  994--1003.

\bibitem[{Li et~al.(2016{\natexlab{b}})Li, Monroe, Ritter, Galley, Gao, and
  Jurafsky}]{li2016deep}
Jiwei Li, Will Monroe, Alan Ritter, Michel Galley, Jianfeng Gao, and Dan
  Jurafsky. 2016{\natexlab{b}}.
\newblock Deep reinforcement learning for dialogue generation.
\newblock \emph{arXiv preprint arXiv:1606.01541}.

\bibitem[{Li et~al.(2019)Li, Weston, and Roller}]{li2019acute}
Margaret Li, Jason Weston, and Stephen Roller. 2019.
\newblock Acute-eval: Improved dialogue evaluation with optimized questions and
  multi-turn comparisons.
\newblock \emph{arXiv preprint arXiv:1909.03087}.

\bibitem[{Liu et~al.(2016)Liu, Lowe, Serban, Noseworthy, Charlin, and
  Pineau}]{D16-1230}
Chia-Wei Liu, Ryan Lowe, Iulian Serban, Mike Noseworthy, Laurent Charlin, and
  Joelle Pineau. 2016.
\newblock \href {https://doi.org/10.18653/v1/D16-1230} {How not to evaluate
  your dialogue system: An empirical study of unsupervised evaluation metrics
  for dialogue response generation}.
\newblock In \emph{Proceedings of the 2016 Conference on Empirical Methods in
  Natural Language Processing}, pages 2122--2132. Association for Computational
  Linguistics.

\bibitem[{Liu et~al.(2019{\natexlab{a}})Liu, Lin, Liu, and Sun}]{liu2019xqa}
Jiahua Liu, Yankai Lin, Zhiyuan Liu, and Maosong Sun. 2019{\natexlab{a}}.
\newblock Xqa: A cross-lingual open-domain question answering dataset.
\newblock In \emph{Proceedings of the 57th Annual Meeting of the Association
  for Computational Linguistics}, pages 2358--2368.

\bibitem[{Liu et~al.(2019{\natexlab{b}})Liu, Shin, Xu, Winata, Xu, Madotto, and
  Fung}]{liu2019zero}
Zihan Liu, Jamin Shin, Yan Xu, Genta~Indra Winata, Peng Xu, Andrea Madotto, and
  Pascale Fung. 2019{\natexlab{b}}.
\newblock Zero-shot cross-lingual dialogue systems with transferable latent
  variables.
\newblock In \emph{Proceedings of the 2019 Conference on Empirical Methods in
  Natural Language Processing and the 9th International Joint Conference on
  Natural Language Processing (EMNLP-IJCNLP)}, pages 1297--1303.

\bibitem[{Liu et~al.(2019{\natexlab{c}})Liu, Winata, Lin, Xu, and
  Fung}]{liu2019attentioninformed}
Zihan Liu, Genta~Indra Winata, Zhaojiang Lin, Peng Xu, and Pascale Fung.
  2019{\natexlab{c}}.
\newblock \href {http://arxiv.org/abs/1911.09273} {Attention-informed
  mixed-language training for zero-shot cross-lingual task-oriented dialogue
  systems}.

\bibitem[{Madotto et~al.(2019)Madotto, Lin, Wu, and
  Fung}]{madotto2019personalizing}
Andrea Madotto, Zhaojiang Lin, Chien-Sheng Wu, and Pascale Fung. 2019.
\newblock Personalizing dialogue agents via meta-learning.
\newblock In \emph{Proceedings of the 57th Annual Meeting of the Association
  for Computational Linguistics}, pages 5454--5459.

\bibitem[{Martin et~al.(2019)Martin, Muller, Su{\'a}rez, Dupont, Romary, de~la
  Clergerie, Seddah, and Sagot}]{martin2019camembert}
Louis Martin, Benjamin Muller, Pedro Javier~Ortiz Su{\'a}rez, Yoann Dupont,
  Laurent Romary, {\'E}ric~Villemonte de~la Clergerie, Djam{\'e} Seddah, and
  Beno{\^\i}t Sagot. 2019.
\newblock Camembert: a tasty french language model.
\newblock \emph{arXiv preprint arXiv:1911.03894}.

\bibitem[{Moon et~al.(2019)Moon, Shah, Kumar, and Subba}]{moon2019opendialkg}
Seungwhan Moon, Pararth Shah, Anuj Kumar, and Rajen Subba. 2019.
\newblock Opendialkg: Explainable conversational reasoning with attention-based
  walks over knowledge graphs.
\newblock In \emph{Proceedings of the 57th Annual Meeting of the Association
  for Computational Linguistics}, pages 845--854.

\bibitem[{Mrk{\v{s}}i{\'c} et~al.(2017)Mrk{\v{s}}i{\'c}, Vuli{\'c},
  S{\'e}aghdha, Leviant, Reichart, Ga{\v{s}}i{\'c}, Korhonen, and
  Young}]{mrkvsic2017semantic}
Nikola Mrk{\v{s}}i{\'c}, Ivan Vuli{\'c}, Diarmuid~{\'O} S{\'e}aghdha, Ira
  Leviant, Roi Reichart, Milica Ga{\v{s}}i{\'c}, Anna Korhonen, and Steve
  Young. 2017.
\newblock Semantic specialization of distributional word vector spaces using
  monolingual and cross-lingual constraints.
\newblock \emph{Transactions of the Association for Computational Linguistics},
  5:309--324.

\bibitem[{Nakayama et~al.(2019)Nakayama, Tjandra, Sakti, and
  Nakamura}]{nakayama2019zero}
Sahoko Nakayama, Andros Tjandra, Sakriani Sakti, and Satoshi Nakamura. 2019.
\newblock Zero-shot code-switching asr and tts with multilingual machine speech
  chain.
\newblock In \emph{2019 IEEE Automatic Speech Recognition and Understanding
  Workshop (ASRU)}, pages 964--971. IEEE.

\bibitem[{Ni et~al.(2017)Ni, Dinu, and Florian}]{ni2017weakly}
Jian Ni, Georgiana Dinu, and Radu Florian. 2017.
\newblock Weakly supervised cross-lingual named entity recognition via
  effective annotation and representation projection.
\newblock In \emph{Proceedings of the 55th Annual Meeting of the Association
  for Computational Linguistics (Volume 1: Long Papers)}, pages 1470--1480.

\bibitem[{Nivre et~al.(2017)Nivre, Agi{\'c}, Ahrenberg
  et~al.}]{nivre2017universal}
Joakim Nivre, {\v{Z}}eljko Agi{\'c}, Lars Ahrenberg, et~al. 2017.
\newblock Universal dependencies 2.0. lindat/clarin digital library at the
  institute of formal and applied linguistics, charles university, prague.

\bibitem[{Pan et~al.(2017)Pan, Zhang, May, Nothman, Knight, and
  Ji}]{pan2017cross}
Xiaoman Pan, Boliang Zhang, Jonathan May, Joel Nothman, Kevin Knight, and Heng
  Ji. 2017.
\newblock Cross-lingual name tagging and linking for 282 languages.
\newblock In \emph{Proceedings of the 55th Annual Meeting of the Association
  for Computational Linguistics (Volume 1: Long Papers)}, volume~1, pages
  1946--1958.

\bibitem[{Papineni et~al.(2002)Papineni, Roukos, Ward, and
  Zhu}]{papineni2002bleu}
Kishore Papineni, Salim Roukos, Todd Ward, and Wei-Jing Zhu. 2002.
\newblock Bleu: a method for automatic evaluation of machine translation.
\newblock In \emph{Proceedings of the 40th annual meeting on association for
  computational linguistics}, pages 311--318. Association for Computational
  Linguistics.

\bibitem[{Pires et~al.(2019)Pires, Schlinger, and
  Garrette}]{pires2019multilingual}
Telmo Pires, Eva Schlinger, and Dan Garrette. 2019.
\newblock How multilingual is multilingual bert?
\newblock In \emph{Proceedings of the 57th Annual Meeting of the Association
  for Computational Linguistics}, pages 4996--5001.

\bibitem[{Radford et~al.(2018)Radford, Narasimhan, Salimans, and
  Sutskever}]{radford2018improving}
Alec Radford, Karthik Narasimhan, Time Salimans, and Ilya Sutskever. 2018.
\newblock Improving language understanding with unsupervised learning.
\newblock \emph{Technical report, OpenAI}.

\bibitem[{Radford et~al.(2019)Radford, Wu, Child, Luan, Amodei, and
  Sutskever}]{radford2019language}
Alec Radford, Jeffrey Wu, Rewon Child, David Luan, Dario Amodei, and Ilya
  Sutskever. 2019.
\newblock Language models are unsupervised multitask learners.
\newblock \emph{OpenAI Blog}, 1(8):9.

\bibitem[{Reddy et~al.(2019)Reddy, Chen, and Manning}]{reddy2019coqa}
Siva Reddy, Danqi Chen, and Christopher~D Manning. 2019.
\newblock Coqa: A conversational question answering challenge.
\newblock \emph{Transactions of the Association for Computational Linguistics},
  7:249--266.

\bibitem[{Rothe et~al.(2019)Rothe, Narayan, and Severyn}]{rothe2019leveraging}
Sascha Rothe, Shashi Narayan, and Aliaksei Severyn. 2019.
\newblock Leveraging pre-trained checkpoints for sequence generation tasks.
\newblock \emph{arXiv preprint arXiv:1907.12461}.

\bibitem[{Sang(2002)}]{sang2002introduction}
Erik~F Sang. 2002.
\newblock Introduction to the conll-2002 shared task: Language-independent
  named entity recognition.
\newblock \emph{arXiv preprint cs/0209010}.

\bibitem[{Sang and De~Meulder(2003)}]{sang2003introduction}
Erik~F Sang and Fien De~Meulder. 2003.
\newblock Introduction to the conll-2003 shared task: Language-independent
  named entity recognition.
\newblock \emph{arXiv preprint cs/0306050}.

\bibitem[{Schuster et~al.(2019{\natexlab{a}})Schuster, Gupta, Shah, and
  Lewis}]{schuster2019cross}
Sebastian Schuster, Sonal Gupta, Rushin Shah, and Mike Lewis.
  2019{\natexlab{a}}.
\newblock Cross-lingual transfer learning for multilingual task oriented
  dialog.
\newblock In \emph{Proceedings of the 2019 Conference of the North American
  Chapter of the Association for Computational Linguistics: Human Language
  Technologies, Volume 1 (Long and Short Papers)}, pages 3795--3805.

\bibitem[{Schuster et~al.(2019{\natexlab{b}})Schuster, Ram, Barzilay, and
  Globerson}]{schuster2019crosslingual}
Tal Schuster, Ori Ram, Regina Barzilay, and Amir Globerson. 2019{\natexlab{b}}.
\newblock Cross-lingual alignment of contextual word embeddings, with
  applications to zero-shot dependency parsing.
\newblock In \emph{Proceedings of the 2019 Conference of the North American
  Chapter of the Association for Computational Linguistics: Human Language
  Technologies, Volume 1 (Long and Short Papers)}, pages 1599--1613.

\bibitem[{See et~al.(2019)See, Roller, Kiela, and Weston}]{see2019makes}
Abigail See, Stephen Roller, Douwe Kiela, and Jason Weston. 2019.
\newblock What makes a good conversation? how controllable attributes affect
  human judgments.
\newblock \emph{arXiv preprint arXiv:1902.08654}.

\bibitem[{Serban et~al.(2016)Serban, Lowe, Charlin, and
  Pineau}]{serban2016generative}
Iulian~Vlad Serban, Ryan Lowe, Laurent Charlin, and Joelle Pineau. 2016.
\newblock Generative deep neural networks for dialogue: A short review.
\newblock \emph{arXiv preprint arXiv:1611.06216}.

\bibitem[{Shuster et~al.(2018)Shuster, Humeau, Bordes, and
  Weston}]{shuster2018engaging}
Kurt Shuster, Samuel Humeau, Antoine Bordes, and Jason Weston. 2018.
\newblock Engaging image chat: Modeling personality in grounded dialogue.
\newblock \emph{arXiv preprint arXiv:1811.00945}.

\bibitem[{Toshniwal et~al.(2018)Toshniwal, Sainath, Weiss, Li, Moreno,
  Weinstein, and Rao}]{toshniwal2018multilingual}
Shubham Toshniwal, Tara~N Sainath, Ron~J Weiss, Bo~Li, Pedro Moreno, Eugene
  Weinstein, and Kanishka Rao. 2018.
\newblock Multilingual speech recognition with a single end-to-end model.
\newblock In \emph{2018 IEEE International Conference on Acoustics, Speech and
  Signal Processing (ICASSP)}, pages 4904--4908. IEEE.

\bibitem[{Vaswani et~al.(2017)Vaswani, Shazeer, Parmar, Uszkoreit, Jones,
  Gomez, Kaiser, and Polosukhin}]{vaswani2017attention}
Ashish Vaswani, Noam Shazeer, Niki Parmar, Jakob Uszkoreit, Llion Jones,
  Aidan~N Gomez, {\L}ukasz Kaiser, and Illia Polosukhin. 2017.
\newblock Attention is all you need.
\newblock In \emph{Advances in neural information processing systems}, pages
  5998--6008.

\bibitem[{Venkatesh et~al.(2018)Venkatesh, Khatri, Ram, Guo, Gabriel, Nagar,
  Prasad, Cheng, Hedayatnia, Metallinou et~al.}]{venkatesh2018evaluating}
Anu Venkatesh, Chandra Khatri, Ashwin Ram, Fenfei Guo, Raefer Gabriel, Ashish
  Nagar, Rohit Prasad, Ming Cheng, Behnam Hedayatnia, Angeliki Metallinou,
  et~al. 2018.
\newblock On evaluating and comparing conversational agents.
\newblock \emph{arXiv preprint arXiv:1801.03625}, 4:60--68.

\bibitem[{Vinyals and Le(2015)}]{vinyals2015neural}
Oriol Vinyals and Quoc~V Le. 2015.
\newblock A neural conversational model.
\newblock \emph{arXiv preprint arXiv:1506.05869}.

\bibitem[{Williams and Young(2007)}]{williams2007partially}
Jason~D Williams and Steve Young. 2007.
\newblock Partially observable markov decision processes for spoken dialog
  systems.
\newblock \emph{Computer Speech \& Language}, 21(2):393--422.

\bibitem[{Winata et~al.(2019{\natexlab{a}})Winata, Lin, and
  Fung}]{winata2019learning}
Genta~Indra Winata, Zhaojiang Lin, and Pascale Fung. 2019{\natexlab{a}}.
\newblock Learning multilingual meta-embeddings for code-switching named entity
  recognition.
\newblock In \emph{Proceedings of the 4th Workshop on Representation Learning
  for NLP (RepL4NLP-2019)}, pages 181--186.

\bibitem[{Winata et~al.(2019{\natexlab{b}})Winata, Lin, Shin, Liu, and
  Fung}]{winata2019hierarchical}
Genta~Indra Winata, Zhaojiang Lin, Jamin Shin, Zihan Liu, and Pascale Fung.
  2019{\natexlab{b}}.
\newblock Hierarchical meta-embeddings for code-switching named entity
  recognition.
\newblock In \emph{Proceedings of the 2019 Conference on Empirical Methods in
  Natural Language Processing and the 9th International Joint Conference on
  Natural Language Processing (EMNLP-IJCNLP)}, pages 3532--3538.

\bibitem[{Winata et~al.(2019{\natexlab{c}})Winata, Madotto, Wu, and
  Fung}]{winata2019code}
Genta~Indra Winata, Andrea Madotto, Chien-Sheng Wu, and Pascale Fung.
  2019{\natexlab{c}}.
\newblock Code-switched language models using neural based synthetic data from
  parallel sentences.
\newblock In \emph{Proceedings of the 23rd Conference on Computational Natural
  Language Learning (CoNLL)}, pages 271--280.

\bibitem[{Wisniewski et~al.(2014)Wisniewski, P{\'e}cheux, Gahbiche-Braham, and
  Yvon}]{wisniewski2014cross}
Guillaume Wisniewski, Nicolas P{\'e}cheux, Souhir Gahbiche-Braham, and
  Fran{\c{c}}ois Yvon. 2014.
\newblock Cross-lingual part-of-speech tagging through ambiguous learning.
\newblock In \emph{Proceedings of the 2014 Conference on Empirical Methods in
  Natural Language Processing (EMNLP)}, pages 1779--1785.

\bibitem[{Wolf et~al.(2019)Wolf, Sanh, Chaumond, and
  Delangue}]{wolf2019transfertransfo}
Thomas Wolf, Victor Sanh, Julien Chaumond, and Clement Delangue. 2019.
\newblock Transfertransfo: A transfer learning approach for neural network
  based conversational agents.
\newblock \emph{arXiv preprint arXiv:1901.08149}.

\bibitem[{Xie et~al.(2018)Xie, Yang, Neubig, Smith, and
  Carbonell}]{xie2018neural}
Jiateng Xie, Zhilin Yang, Graham Neubig, Noah~A Smith, and Jaime Carbonell.
  2018.
\newblock Neural cross-lingual named entity recognition with minimal resources.
\newblock In \emph{Proceedings of the 2018 Conference on Empirical Methods in
  Natural Language Processing}, pages 369--379.

\bibitem[{Yavuz et~al.(2019)Yavuz, Rastogi, Chao, and
  Hakkani-Tur}]{yavuz2019deepcopy}
Semih Yavuz, Abhinav Rastogi, Guan-Lin Chao, and Dilek Hakkani-Tur. 2019.
\newblock Deepcopy: Grounded response generation with hierarchical pointer
  networks.
\newblock In \emph{Proceedings of the 20th Annual SIGdial Meeting on Discourse
  and Dialogue}, pages 122--132.

\bibitem[{Young et~al.(2013)Young, Ga{\v{s}}i{\'c}, Thomson, and
  Williams}]{young2013pomdp}
Steve Young, Milica Ga{\v{s}}i{\'c}, Blaise Thomson, and Jason~D Williams.
  2013.
\newblock Pomdp-based statistical spoken dialog systems: A review.
\newblock \emph{Proceedings of the IEEE}, 101(5):1160--1179.

\bibitem[{Yue et~al.(2019)Yue, Lee, Y{\i}lmaz, Deng, and Li}]{yue2019end}
Xianghu Yue, Grandee Lee, Emre Y{\i}lmaz, Fang Deng, and Haizhou Li. 2019.
\newblock End-to-end code-switching asr for low-resourced language pairs.
\newblock \emph{arXiv preprint arXiv:1909.12681}.

\bibitem[{Zemlyanskiy and Sha(2018)}]{zemlyanskiy2018aiming}
Yury Zemlyanskiy and Fei Sha. 2018.
\newblock Aiming to know you better perhaps makes me a more engaging dialogue
  partner.
\newblock \emph{CoNLL 2018}, page 551.

\bibitem[{Zhang et~al.(2018)Zhang, Dinan, Urbanek, Szlam, Kiela, and
  Weston}]{personachat}
Saizheng Zhang, Emily Dinan, Jack Urbanek, Arthur Szlam, Douwe Kiela, and Jason
  Weston. 2018.
\newblock \href {http://aclweb.org/anthology/P18-1205} {Personalizing dialogue
  agents: I have a dog, do you have pets too?}
\newblock In \emph{Proceedings of the 56th Annual Meeting of the Association
  for Computational Linguistics (Volume 1: Long Papers)}, pages 2204--2213.
  Association for Computational Linguistics.

\bibitem[{Zhang et~al.(2019)Zhang, Sun, Galley, Chen, Brockett, Gao, Gao, Liu,
  and Dolan}]{zhang2019dialogpt}
Yizhe Zhang, Siqi Sun, Michel Galley, Yen-Chun Chen, Chris Brockett, Xiang Gao,
  Jianfeng Gao, Jingjing Liu, and Bill Dolan. 2019.
\newblock Dialogpt: Large-scale generative pre-training for conversational
  response generation.
\newblock \emph{arXiv preprint arXiv:1911.00536}.

\bibitem[{Zhang et~al.(2016)Zhang, Gaddy, Barzilay, and
  Jaakkola}]{zhang2016ten}
Yuan Zhang, David Gaddy, Regina Barzilay, and Tommi Jaakkola. 2016.
\newblock Ten pairs to tag--multilingual pos tagging via coarse mapping between
  embeddings.
\newblock In \emph{Proceedings of the 2016 Conference of the North American
  Chapter of the Association for Computational Linguistics: Human Language
  Technologies}, pages 1307--1317.

\bibitem[{Zhao et~al.(2017)Zhao, Zhao, and Eskenazi}]{zhao2017learning}
Tiancheng Zhao, Ran Zhao, and Maxine Eskenazi. 2017.
\newblock Learning discourse-level diversity for neural dialog models using
  conditional variational autoencoders.
\newblock In \emph{Proceedings of the 55th Annual Meeting of the Association
  for Computational Linguistics (Volume 1: Long Papers)}, pages 654--664.

\end{thebibliography}
\bibliographystyle{acl_natbib}

\appendix
\clearpage
\section{Dataset Collection}
\subsection{Annotation Instructions}
In this section, we show the instructions for French annotation:
\begin{itemize}
    \item There are two existing columns of conversations: the first column (en) is the original conversations in English, the second column (fr) is the conversations translated by an automatic system (e.g., Google Translate).
    \item You should copy the conversation from the second column (the translated conversations) into the third column (named fr\_annotation). In that column, you should then revise the incorrect or inappropriate translations.
    \item The goal of the revision is to make the conversations more coherent and fluent in the target language (French). Hence you can customize dialogues and persona sentences to make them fluent and coherent in the target language, including by deviating from the original translation. However, you should retain persona and conversation consistency.
\end{itemize}
\subsection{Training Set Statistics}
We report our iterative revised training set statistics in Table \ref{table:train}. 
% The training set has much lower edit distance comparing to validation and test set, since it 

\begin{table}[!b]
\begin{tabular}{rcccc}
\hline
\multicolumn{5}{c}{\textbf{Train}} \\ \hline
\multicolumn{1}{c|}{\textit{\textbf{Lang}}} & \multicolumn{1}{c|}{\textit{\textbf{\# Dial.}}} & \multicolumn{1}{c|}{\textit{\textbf{\# Utt.}}} & \multicolumn{1}{c|}{\textit{\textbf{Edit}}} & \textit{\textbf{BLEU}} \\ \hline
\multicolumn{1}{r|}{\textit{\textbf{Fr}}} & \multicolumn{1}{c|}{16878} & \multicolumn{1}{c|}{248244} & \multicolumn{1}{c|}{0.06} & 99.98 \\
\multicolumn{1}{r|}{\textit{\textbf{It}}} & \multicolumn{1}{c|}{16878} & \multicolumn{1}{c|}{248244} & \multicolumn{1}{c|}{1.09} & 99.8 \\
\multicolumn{1}{r|}{\textit{\textbf{Id}}} & \multicolumn{1}{c|}{16878} & \multicolumn{1}{c|}{248244} & \multicolumn{1}{c|}{0.18} & 99.94 \\
\multicolumn{1}{r|}{\textit{\textbf{Jp}}} & \multicolumn{1}{c|}{16878} & \multicolumn{1}{c|}{248244} & \multicolumn{1}{c|}{0.38} & 99.17 \\
\multicolumn{1}{r|}{\textit{\textbf{Ko}}} & \multicolumn{1}{c|}{16878} & \multicolumn{1}{c|}{248244} & \multicolumn{1}{c|}{0.97} & 99.51 \\
\multicolumn{1}{r|}{\textit{\textbf{Zh}}} & \multicolumn{1}{c|}{16878} & \multicolumn{1}{c|}{248244} & \multicolumn{1}{c|}{0.52} & 98.1 \\ \hline
\end{tabular}
\caption{The number of dialogues (\#Dial.) and utterances (\#Utt.) of the training set in six languages. Edit distance per dialogue and BLEU score are computed to show the difference between the iterative revised dataset and auto-translated dataset.}
\label{table:train}
\end{table}

 \begin{figure}[t]
    \centering
        \includegraphics[width=\linewidth]{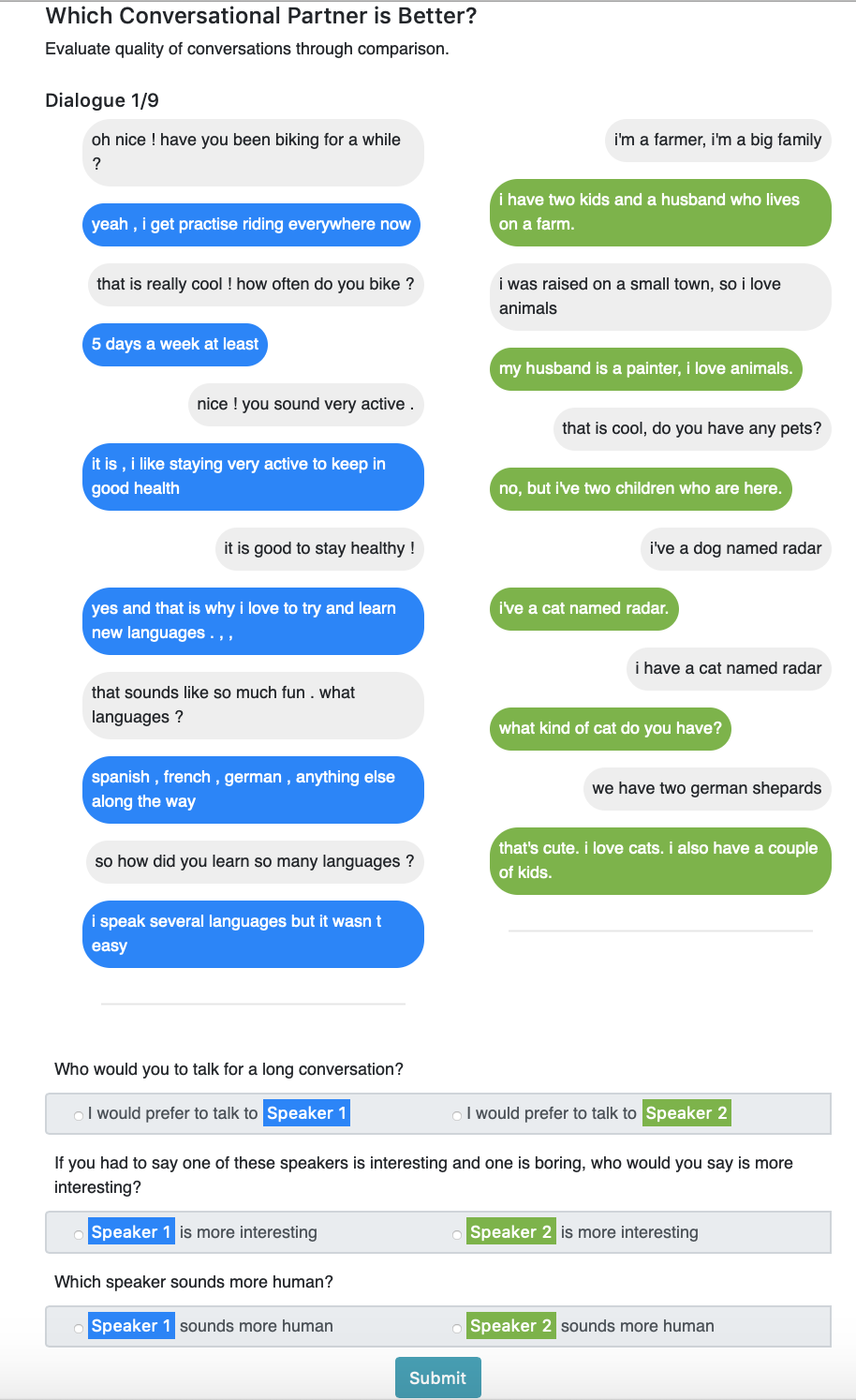}
    \caption{Human evaluation interface modified from ACUTE-EVAL\cite{li2019acute}}
\label{fig:interface}
\end{figure}
\section{Model Detail}
\begin{figure*}[t]
    \centering
    \resizebox{0.99\textwidth}{!}{
    % \begin{minipage}[!t]{\textwidth}
     \input{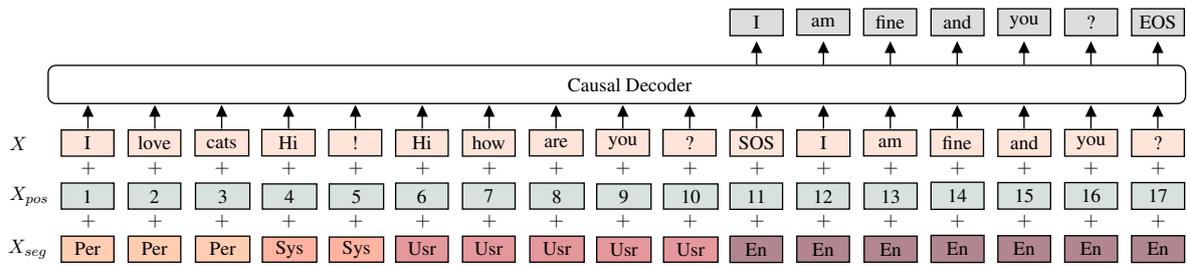}
     }
    % \end{minipage}
           
    \caption{Multilingual Causal Decoder model. }
    \label{fig:Detail1}
\end{figure*}

\begin{figure*}[t]
    \centering
    \resizebox{0.99\textwidth}{!}{
    % \begin{minipage}[!t]{\textwidth}
     \input{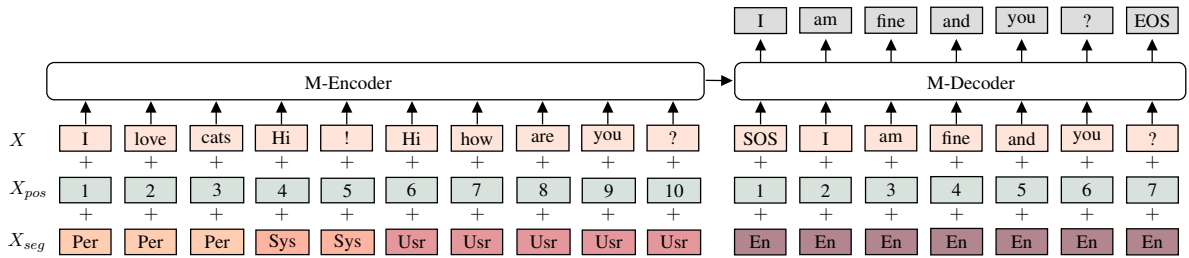}
     }
    % \end{minipage}
           
    \caption{Multilingual Encoder-Decoder model. }
    \label{fig:Detail2}
\end{figure*}

Figure \ref{fig:Detail1} and \ref{fig:Detail2} illustrates the details of the multilingual causal decoder and the multilingual encoder-decoder models.


\section{Human Evaluation}
As illustrated in Figure \ref{fig:interface}, the annotator is provided with two full dialogues made by a self-chat model or human-dialogues. Then the annotators are asked the following questions:
\begin{itemize}
    \item Who would you talk to for a long conversation? 
    \item If you had to say one of these speakers is interesting and one is boring, who would you say is more interesting?
    \item Which speaker sounds more human?
\end{itemize}

\section{Generated Samples}
\subsection{Mixed-language Samples}
We report more the mixed-language samples generated by M-CausalBert in Table \ref{table:one2many_a} and \ref{table:many2one_a}.

\begin{table*}[th]
\resizebox{0.99\linewidth}{!}{
\begin{tabular}{ll}
\multicolumn{2}{c}{\textbf{System Persona}} \\
\multicolumn{2}{c}{\begin{tabular}[c]{@{}c@{}}i have two dogs. \\ i like to work on vintage cars. \\ my favorite music is country. \\ i own two vintage mustangs.\end{tabular}} \\
\multicolumn{2}{c}{\textbf{Dialogue}} \\ \hline
{\color[HTML]{CB0000} Usr} & do you have any hobbies? \\ \hline
{\color[HTML]{32CB00} En} & \begin{tabular}[c]{@{}l@{}}i like music, country is my favorite. you?\end{tabular} \\ \hline
{\color[HTML]{32CB00} Zh} & \begin{tabular}[c]{@{}l@{}}\begin{CJK*}{UTF8}{gbsn}我喜欢听乡村音乐。你呢\end{CJK*} \\ (I like country music, you?)\end{tabular} \\ \hline
{\color[HTML]{32CB00} Fr} & \begin{tabular}[c]{@{}l@{}}je possède deux voitures anciennes \\ et j'aime faire de l'artisanat\\ (I have two vintage cars and I like to make crafts)\end{tabular} \\ \hline
{\color[HTML]{32CB00} It} & \begin{tabular}[c]{@{}l@{}}Ciao. mi piace lavorare su auto d'epoca \\ e ascoltare musica country.\\ (Hello. I love working on vintage cars and\\ listening to country music.)\end{tabular} \\ \hline
{\color[HTML]{32CB00} Jp} & \begin{tabular}[c]{@{}l@{}}\begin{CJK*}{UTF8}{min}こんにちは。私は大学にいます、\end{CJK*}\\ (Hi, i really like it)\end{tabular} \\ \hline
{\color[HTML]{32CB00} Id} & \begin{tabular}[c]{@{}l@{}}Saya suka mendengarkan musik country, dan Anda? \\ (I like listening to country music, and you?)\end{tabular} \\ \hline
{\color[HTML]{32CB00} Ko} & \begin{tabular}[c]{@{}l@{}}\begin{CJK*}{UTF8}{}\CJKfamily{mj}안녕, 나는 선생님이야. \end{CJK*}\\(I have two vintage mustangs.)\end{tabular} \\ \hline
\end{tabular}

\begin{tabular}{ll}
\multicolumn{2}{c}{\textbf{System Persona}} \\
\multicolumn{2}{c}{\begin{tabular}[c]{@{}c@{}}i love disneyland and mickey mouse. \\ i love to spend time with my family. \\ i'm a baby delivery nurse. \\ i walk three miles every day.\end{tabular}} \\
\multicolumn{2}{c}{\textbf{Dialogue}} \\ \hline
{\color[HTML]{CB0000} Usr} & what do you do in your free time? \\ \hline
{\color[HTML]{32CB00} En} & \begin{tabular}[c]{@{}l@{}}i love to play with my family. how about you?\end{tabular} \\ \hline
{\color[HTML]{32CB00} Zh} & \begin{tabular}[c]{@{}l@{}}\begin{CJK*}{UTF8}{gbsn}我主要是在家里\end{CJK*} \\ (I just stay at home)\end{tabular} \\ \hline
{\color[HTML]{32CB00} Fr} & \begin{tabular}[c]{@{}l@{}}j'aime passer du temps avec ma\\ famille et mon mari.\\ (i love spending time with my family and \\my husband.)\end{tabular} \\ \hline
{\color[HTML]{32CB00} It} & \begin{tabular}[c]{@{}l@{}}Ciao. Sono in pensione da quando ero un'infermiera.\\ (Hello. I have been retired since I was a nurse.)\end{tabular} \\ \hline
{\color[HTML]{32CB00} Jp} & \begin{tabular}[c]{@{}l@{}}\begin{CJK*}{UTF8}{min}こんにちは、私は家族と時間を過ごすのが\end{CJK*} \\\begin{CJK*}{UTF8}{min}大好きです\end{CJK*}\\ (Hi, i love spending time with my family)\end{tabular} \\ \hline
{\color[HTML]{32CB00} Id} & \begin{tabular}[c]{@{}l@{}}Saya suka berjalan 3 mil setiap hari dan Anda? \\ (I love walking 3 miles every day and you?)\end{tabular} \\ \hline
{\color[HTML]{32CB00} Ko} & \begin{tabular}[c]{@{}l@{}}\begin{CJK*}{UTF8}{}\CJKfamily{mj}안녕, 나는 가족과 시간을 보내는 것을 즐긴다. \end{CJK*}\\(Hi, I enjoy spending time with my family.)\end{tabular} \\ \hline
\end{tabular}
}
\caption{One-to-many by M-CausalBert}
\label{table:one2many_a}
\end{table*}

\begin{table*}[th]
\resizebox{0.99\linewidth}{!}{
\begin{tabular}{ll}
\multicolumn{2}{c}{\textbf{System Persona}} \\
\multicolumn{2}{c}{\begin{tabular}[c]{@{}c@{}}i am a cancer survivor. \\ my parents were both very athletic. \\ i got a new job just yesterday to be a life coach. \\ i am a clean eater. \end{tabular}} \\
\multicolumn{2}{c}{\textbf{Dialogue}} \\ \hline
{\color[HTML]{CB0000} Usr} & \begin{tabular}[c]{@{}l@{}}\begin{CJK*}{UTF8}{min}こんにちは \end{CJK*} \\ (Hello)\end{tabular} \\ \hline
{\color[HTML]{32CB00} Sys} & hey there how are you? \\ \hline
{\color[HTML]{CB0000} Usr} & \begin{tabular}[c]{@{}l@{}}\begin{CJK*}{UTF8}{gbsn}我很好，我是一名学生，你呢？\end{CJK*}\\ (I am good, I am a student, what about you?)\end{tabular} \\ \hline
{\color[HTML]{32CB00} Sys} & \begin{tabular}[c]{@{}l@{}}i am a full time student and have no job yet.\end{tabular} \\ \hline
{\color[HTML]{CB0000} Usr} & \begin{tabular}[c]{@{}l@{}}cosa Studi? \\ (What do you study?)\end{tabular} \\ \hline
{\color[HTML]{32CB00} Sys} & oh i see. what are you studying? \\ \hline
{\color[HTML]{CB0000} Usr} & \begin{tabular}[c]{@{}l@{}}l'informatique \\ (computer science)\end{tabular} \\ \hline
{\color[HTML]{32CB00} Sys} & \begin{tabular}[c]{@{}l@{}}computer stuff. what is your favorite subject?\end{tabular} \\ \hline
{\color[HTML]{CB0000} Usr} & \begin{tabular}[c]{@{}l@{}}i like machine learning.\end{tabular} \\ \hline
{\color[HTML]{32CB00} Sys} & i like to read and make fun of things.\\ \hline
{\color[HTML]{CB0000} Usr} & \begin{tabular}[c]{@{}l@{}}Apa pekerjaan orang tuamu?\\ (what do your parents do?)\end{tabular} \\ \hline
{\color[HTML]{32CB00} Sys} & i'm from cali, my parents were both athletic.\\ \hline
\end{tabular}

\begin{tabular}{ll}
\multicolumn{2}{c}{\textbf{System Persona}} \\
\multicolumn{2}{c}{\begin{tabular}[c]{@{}c@{}}my brother is older than me. \\ i ride the bus to school. \\ i'm on the soccer team. \\ i am 13 years old. \end{tabular}} \\
\multicolumn{2}{c}{\textbf{Dialogue}} \\ \hline

{\color[HTML]{CB0000} Usr} & \begin{tabular}[c]{@{}l@{}}\begin{CJK*}{UTF8}{}\CJKfamily{mj}안녕 \end{CJK*} \\ (Hello)\end{tabular} \\ \hline
{\color[HTML]{32CB00} Sys} & hello, how are you? \\ \hline
{\color[HTML]{CB0000} Usr} & \begin{tabular}[c]{@{}l@{}}\begin{CJK*}{UTF8}{gbsn}我很好，刚刚下班回家\end{CJK*}\\ (I am good, just came home from work)\end{tabular} \\ \hline

{\color[HTML]{32CB00} Sys} & \begin{tabular}[c]{@{}l@{}}oh, cool. what do you do for work?\end{tabular} \\ \hline
{\color[HTML]{CB0000} Usr} & \begin{tabular}[c]{@{}l@{}}Sono un insegnante di inglese \\ (I am English teacher.)\end{tabular} \\ \hline
{\color[HTML]{32CB00} Sys} &oh, that's awesome! i hope to get a job soon. \\ \hline
{\color[HTML]{CB0000} Usr} & \begin{tabular}[c]{@{}l@{}}quel âge avez-vous? \\ (how old are you? )\end{tabular} \\ \hline
{\color[HTML]{32CB00} Sys} & \begin{tabular}[c]{@{}l@{}}oh ok. i'm only 13, but my brother is\\ older than me\end{tabular} \\ \hline
{\color[HTML]{CB0000} Usr} & \begin{tabular}[c]{@{}l@{}}Apakah Anda pelajar? (are you a student?)\end{tabular} \\ \hline
{\color[HTML]{32CB00} Sys} & do you play soccer? i am a football player\\ \hline
{\color[HTML]{CB0000} Usr} & \begin{tabular}[c]{@{}l@{}} no, i like to read\end{tabular} \\ \hline
{\color[HTML]{32CB00} Sys} & i like to ride the bus and play soccer\\ \hline
\end{tabular}
}
\caption{Many-to-one by M-CausalBert}
\label{table:many2one_a}
\end{table*}

\subsection{Model Comparison Samples}
We randomly sample one self-chat dialogue examples for each model in each language and report them in figure 5-37.
\clearpage

\def\names{{CausalBert},{M-CausalBert},{PolyEncoder},{M-Bert2Bert}}
\foreach \name in \names{
     \begin{figure}
         \noindent\includegraphics[width=\linewidth]{EN/\name/A_0.pdf}
         \caption{English \name}
     \end{figure}
}

\def\names{{CausalBert},{M-CausalBert},{PolyEncoder},{M-Bert2Bert},{CrossLingual}}
\foreach \name in \names{
     \begin{figure}
         \noindent\includegraphics[width=\linewidth]{ZH/\name/A_2.pdf}
         \caption{Chinese \name}
     \end{figure}
}

\foreach \name in \names{
     \begin{figure}
         \noindent\includegraphics[width=\linewidth]{IT/\name/A_0.pdf}
         \caption{Italian \name}
     \end{figure}
}

\foreach \name in \names{
     \begin{figure}
         \noindent\includegraphics[width=\linewidth]{FR/\name/A_0.pdf}
         \caption{France \name}
     \end{figure}
}

\foreach \name in \names{
     \begin{figure}
         \noindent\includegraphics[width=\linewidth]{JP/\name/A_1.pdf}
         \caption{Japanese \name}
     \end{figure}
}

\def\names{{CausalBert},{M-CausalBert},{PolyEncoder},{M-Bert2Bert}}
\foreach \name in \names{
     \begin{figure}
         \noindent\includegraphics[width=\linewidth]{KO/\name/A_0.pdf}
         \caption{Korean \name}
     \end{figure}
}
\def\names{{CausalBert},{M-CausalBert},{PolyEncoder},{M-Bert2Bert},{CrossLingual}}
\foreach \name in \names{
     \begin{figure}
         \noindent\includegraphics[width=\linewidth]{ID/\name/A_0.pdf}
         \caption{Indonesian \name}
     \end{figure}
}

\end{document}